# Any-to-All MRI Synthesis: A Unified Foundation Model for Nasopharyngeal Carcinoma and Its Downstream Applications

Yao Pu, Yiming Shi, Zhenxi Zhang, Peixin Yu, Yitao Zhuang, Xiang Wang, Hongzhao Chen, Jing Cai*, Ge Ren*

*Institution*: Department of Health Technology and Informatics, The Hong Kong Polytechnic University, Kowloon, Hong Kong SAR, China. *Email*: allen-yao.pu@connect.polyu.hk, gary-ge.ren@polyu.edu.hk.

## Abstract

Magnetic resonance imaging (MRI) is essential for nasopharyngeal carcinoma (NPC) radiotherapy (RT), but practical constraints, such as patient discomfort, long scan times, and high costs often lead to incomplete modalities in clinical practice, compromising RT planning accuracy. Traditional MRI synthesis methods are modality-specific, limited in anatomical adaptability, and lack clinical interpretability—failing to meet NPC's RT needs. Here, we developed a unified foundation model integrating contrastive visual representation learning and vision–language alignment (VLA) to enable any-to-all MRI synthesis. The model uses a contrastive encoder for modality-invariant representations and a CLIP-based text-informed decoder for semantically consistent synthesis, supporting any-to-all MRI synthesis via one unified foundation model. Trained on 40,825 images from 13 institutions, it achieves consistently high performance (average SSIM 0.90, PSNR 27) across 26 internal/external validation sites (15,748 images), with superior synthesis fidelity and robustness to noise and domain shifts. Meanwhile, its unified representation enhances downstream RT-relevant tasks (e.g., segmentation). This work advances digital medicine solutions for NPC care by leveraging foundation models to bridge technical synthesis and clinical utility.

## Introduction

Magnetic resonance imaging (MRI) is pivotal in radiation oncology, offering exceptional soft tissue contrast and multi-contrast sequence capabilities to characterize tumor morphology, composition, and heterogeneity [1,2]. For example, T1-weighted, T2-weighted, contrast-enhanced T1 (T1c), diffusion-weighted imaging (DWI), and FLAIR sequences each provide unique diagnostic insights [3,4]. However, acquiring complete multi-contrast protocols is often unfeasible due to prolonged scan times, patient discomfort, contrast-agent contraindications, or cost—leading to incomplete datasets that hinder accurate diagnosis and quantitative analysis [5,6]. This has driven growing interest in MRI synthesis, a digital medicine strategy to synthesize the missing modalities and support clinical decision-making [7–9].

The challenge of incomplete MRI is acutely relevant to nasopharyngeal carcinoma (NPC), where MRI

underpins the entire radiotherapy (RT) workflow [10]. The nasopharynx's anatomical complexity and proximity to critical structures demand precise tumor delineation for effective RT. Multi-sequence MRI can be used for NPC TNM staging (e.g., assessing skull base invasion, lymph node metastases) and integrating with planning computed tomography (CT) to define target volumes and guide dose planning [9,11,12,13]. Furthermore, MRI continues to play an essential role during treatment delivery and follow-up, serving as the reference for image guidance, adaptive radiotherapy, and post-treatment response assessment [14]. Missing key sequences (e.g., T1c for tumor enhancement, DWI for viable disease detection) directly compromises target accuracy, organ-at-risk sparing, and treatment outcomes. However, practical constraints often lead to incomplete modalities in clinical practice, which elevates MRI synthesis from a technical convenience to a clinically critical need for NPC treatment.

Early efforts in MRI synthesis primarily relied on deep learning models based on generative adversarial networks (GANs), which demonstrated the feasibility of cross-modality image translation through adversarial training [15–20]. Subsequent developments introduced diffusion-based generative models, achieving improved training stability by modeling image synthesis as a progressive denoising process [21–23]. More recently, Transformer-based architectures and emerging state-space models have been explored to capture long-range spatial dependencies and global contextual information in medical images [24–27]. However, most existing approaches are input modality-specific and limited to single anatomical organ like the brain, failing to adapt to NPC's complex head-and-neck anatomy and variable clinical workflow gaps. Critically, these models lack clinical interpretability, providing little insight into the pathological semantics of synthesized content. Moreover, evaluations often prioritize pixel-level metrics over downstream RT-relevant tasks (e.g., tumor segmentation), limiting clinical translatability [8,28,29]. These limitations of previous models demand a paradigm shift—one that only foundation models can deliver.

Foundation models are uniquely positioned to overcome the shortcomings of traditional MRI synthesis for NPC RT. Firstly, pre-trained on diverse datasets, they capture universal anatomical patterns and develop high generalization, enabling them to handle missing variable sequences in NPC workflows without retraining [30-32]. Secondly, their integration with vision-language models (VLMs)—such as clinical adaptations of CLIP—solves the critical issue of poor clinical interpretability [33-36]. By aligning visual features with clinical semantics (e.g., "skull base invasion"), they encode explicit radiological concepts into the synthesis process, bridging low-level pixels and high-level clinical interpretation—a key advantage for NPC, where assessment hinges on language-describable pathology. Beyond synthesis, the inherent multi-task adaptability of foundation models aligns directly with the multi-stage demands of NPC RT (e.g., staging, target delineation, response assessment) [8, 28, 29, 38, 39]. Their pre-trained representations can be fine-tuned for RT-specific downstream tasks without sacrificing performance. This adaptability ultimately bridges the persistent gap between mere image synthesis quality and genuine clinical utility that has plagued traditional methods.

However, the integration of foundation models with VLMs for multi-sequence MRI synthesis and RT-oriented tasks in NPC remains underexplored. To bridge this gap, we propose a novel framework featuring a contrastive vision encoder for modality-invariant anatomical representations and a text-informed decoder (fine-tuned from a CLIP-based VLM) to guide semantically consistent image generation. This unified architecture enables any-to-all MRI sequence synthesis while facilitating effective knowledge transfer across imaging domains. Evaluated on multi-center NPC datasets, our framework demonstrates superior performance compared to modality-specific models. It achieves high-fidelity synthesis and, critically, enhances robustness in downstream segmentation tasks essential for RT workflows. This approach directly advances digital medicine by translating model capability into practical, unmet clinical needs in NPC care.

## Results and discussion

### Overview

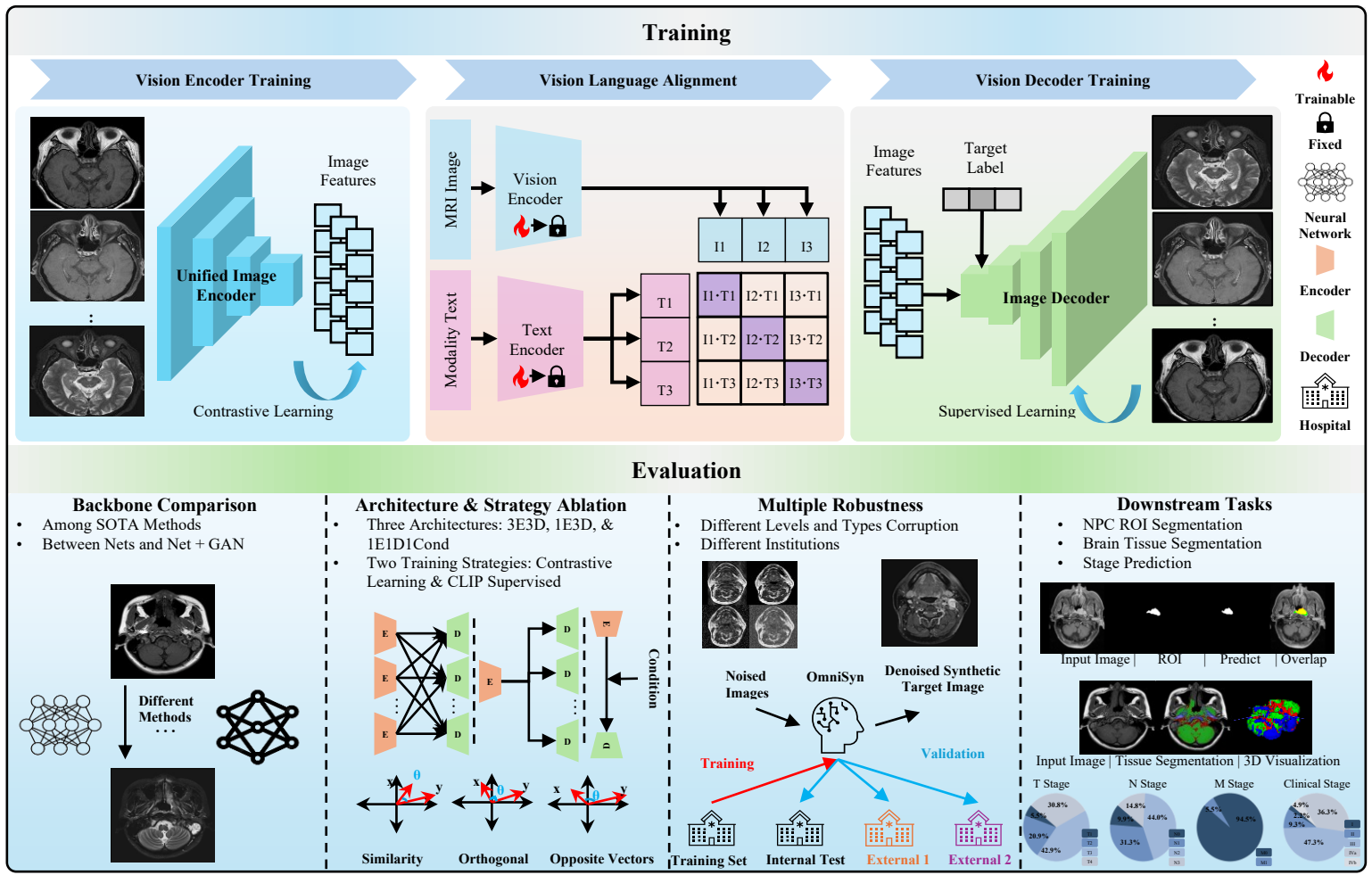

**Figure 1.** Study Design and Experiments. The framework follows a three-stage training strategy (vision encoder pre-training, vision-language alignment, and text-informed decoder training), followed by a four-

part evaluation (comparative analysis, ablation study, robustness testing, and downstream RT tasks assessment).

**Evaluation experiments**

This study provides a comprehensive evaluation of the proposed MRI synthesis framework across diverse datasets, tasks, and settings through four core experimental modules: **First**, we quantify the overall synthesis performance on internal and external datasets spanning varied acquisition protocols and clinical conditions (in section *Overall performance on internal and external datasets*). This evaluation incorporates systematic comparisons with mainstream synthesis methods, backbone networks, and their variants integrated with GAN-based refinement modules. **Second**, we perform a full ablation study to dissect the contributions of architectural designs and training strategies (in section *Ablation Study*). Specifically, we first analyze the impact of individual architectural components; we then further investigate the standalone and synergistic effects of two training strategies on synthesis performance. Additionally, we conduct an intermediate analysis of the unified encoder's contrastive pretraining process and verify the behavioral characteristics of multiple CLIP variants under zero-shot and fine-tuning scenarios. **Third**, we assess model robustness by introducing graded controlled degradations (e.g., motion artifacts, Gaussian noise, etc.) to test the stability of synthesized MRI images under challenging conditions (in section *Robustness quantification*). **Fourth**, we validate the downstream utility of synthesized images in clinically relevant tasks, including NPC tumor segmentation, brain area segmentation and stage prediction, to underscore the real-world applicability of our approach (in section *Downstream tasks evaluation*). **Finally**, we discuss the limitations of the proposed framework and outline prospective future research directions (in section *Outlook*).

**Evaluation metrics**

To quantitatively assess the performance of the proposed framework, we employed several complementary evaluation metrics tailored to different experimental tasks. **For the synthesis mission**, we used mean squared error (MSE), peak signal-to-noise ratio (PSNR), and structural similarity index measure (SSIM) to evaluate the voxel-wise fidelity and perceptual similarity between the synthesized and reference MR images [40,41]. MSE quantifies the pixel-level difference between the generated and ground truth (GT) images, with lower values indicating superior reconstruction accuracy. PSNR, derived from MSE, measures the logarithmic ratio between the maximum possible signal intensity and the noise level, reflecting global image quality. SSIM evaluates the similarity between the two images in terms of luminance, contrast, and structural information, with higher values suggesting better perceptual quality and anatomical preservation. **For the downstream segmentation task**, we used the Dice Similarity Coefficient (DSC) to measure the overlap between the predicted and reference segmentation masks. Dice is defined as twice the intersection of the two regions divided by their combined volume, where higher values denote more accurate structural delineation. **For the modality prediction task**, we adopted modality classification accuracy as the evaluation criterion, defined as the proportion of correctly classified samples relative to the total number of test samples. For the vector's alignment, cosine similarity [42] was selected as the criterion.

$$\text{MSE} = \frac{1}{N}\sum_{i=1}^{N}(I_i - \hat{I}_i)^2$$

$$\text{PSNR} = 10 \cdot \log_{10}\left(\frac{L^2}{\text{MSE}}\right)$$

$$\text{SSIM}(I, \hat{I}) = \frac{(2\mu_I\mu_{\hat{I}} + C_1)(2\sigma_{I\hat{I}} + C_2)}{(\mu_I^2 + \mu_{\hat{I}}^2 + C_1)(\sigma_I^2 + \sigma_{\hat{I}}^2 + C_2)} \quad (1)$$

$$\text{DSC} = \frac{2|P \cap G|}{|P| + |G|}$$

$$\text{ACC} = \frac{TP + TN}{TP + TN + FP + FN}$$

$$\text{CosineSimilarity}(A, B) = \frac{A \cdot B}{\|A\|\|B\|}$$

Where $I_i$ and $\hat{I}_i$ denote the voxel intensities of the reference and synthesized images, respectively, and $N$ is the total number of voxels. $L$ is the maximum possible pixel intensity value of the image. $\mu_I$ and $\mu_{\hat{I}}$ are the mean intensities, $\sigma_I^2$ and $\sigma_{\hat{I}}^2$ are the variances, and $\sigma_{I\hat{I}}$ is the covariance between $I_i$ and $\hat{I}_i$. $C_1$ and $C_2$ are constant to stabilize the division. $P$ and $G$ represent the predicted and ground-truth segmentation regions, respectively. $TP$, $TN$, $FP$, and $FN$ denote the numbers of true positives, true negatives, false positives, and false negatives, respectively. $A \cdot B$ means the dot product, and $\|*\|$ means the L2 norm of vector.

### Overall performance on internal and external datasets

#### Comparison with mainstream synthesis methods

This experiment focuses on **method-level comparison**, where we benchmark the proposed unified MRI synthesis framework (OmniSyn) against representative mainstream approaches, including convolution-based networks, transformer-based architectures, adversarial translation models, diffusion models, and recently published state-of-the-art methods. Convolutional adversarial baselines such as **pix2pix** [43] and **CycleGAN** [44] provide efficient paired and unpaired image-to-image translation but rely heavily on local convolutional receptive fields, making them less capable of capturing long-range anatomical dependencies. Transformer-based approaches address this limitation: architectures such as **SwinUNet** [45] and **ResViT** [24] introduce global self-attention and hierarchical windowed attention to model spatial relationships more effectively. Moreover, the **DDPM** [31] family represents diffusion-based synthesis, leveraging iterative denoising to learn high-fidelity structural distributions. Additionally, the newest models **BrainMVP** [39] and **TUMSyn** [28] combined language model for MRI synthesis, which both are also replicated as comparison methods.

Across both internal and external datasets (**Figure 2**), our OmniSyn model consistently outperforms all mainstream baselines. Quantitatively, OmniSyn achieves the lowest MSE and the highest SSIM and PSNR among all evaluated approaches. Visually, OmniSyn generates more anatomically coherent structures, with sharper tissue boundaries and fewer hallucinated textures. ResViT, BrainMVP, TUMSyn, and DDPM models deliver relatively strong detail recovery but suffer from artifacts, or over smoothing, or long inference time, or additional text input, while pix2pix and CycleGAN struggle with subtle anatomical preservation. Transformer-based SwinUnet method performs worst among all methods, particularly in cross-domain generalization, which is caused by the windows splitting and merging. These results demonstrate that the unified vision-language-guided representation and structurally aligned decoder in OmniSyn provide superior robustness and fidelity compared with conventional paradigms.

**On internal dataset: Figure 2 (a)** presents a qualitative comparison of synthesized results on the internal testing dataset. Our foundation model demonstrated visually superior performance in generating contrast-enhanced MRI images, with improved anatomical structure continuity and reduced texture artifacts compared with competing networks. Finer boundary details in the tumor and surrounding tissues were more accurately preserved. Competing models tended to produce slightly blurred textures or inconsistent intensity distributions, especially in regions with low signal-to-noise ratios. Quantitative analysis, summarized in **Table 1**, further confirmed these findings. Our model achieved the lowest mean squared error (MSE) and the highest peak signal-to-noise ratio (PSNR) and structural similarity index (SSIM) values among all compared methods, indicating superior fidelity and perceptual quality of the synthesized images. The improvements were statistically significant ($P < 0.001$) compared with each baseline method.

**On external datasets:** To validate model robustness and cross-institutional generalization, the same networks were tested on two independent external datasets acquired from different hospitals. While all networks exhibited a moderate performance decline due to domain shifts in scanner type and acquisition protocol, our model maintained the highest reconstruction quality across both datasets. On External-1 (**Figure 2 (b)**), our model achieved consistent enhancement of soft-tissue contrast and structural alignment, whereas other methods produced over-smoothed or distorted regions in some slices. On External-2 (**Figure 2 (c)**), which contained higher noise levels and intensity variations, our model effectively preserved the anatomical structures and reduced hallucinated details, demonstrating strong robustness against unseen data distributions.

**Table 1. Comparison among novel synthesis methods on three datasets.**

| | Internal | | | | | | | | External 1 | | | | | | | | External 2 | | | | | | | |
|---|---|---|---|---|---|---|---|---|---|---|---|---|---|---|---|---|---|---|---|---|---|---|---|---|
| | pix2pix | CycleGAN | SwinUnet | ResViT | DDPM | BrainMVP | TUMSyn | OmniSyn | pix2pix | CycleGAN | SwinUnet | ResViT | DDPM | BrainMVP | TUMSyn | OmniSyn | pix2pix | CycleGAN | SwinUnet | ResViT | DDPM | BrainMVP | TUMSyn | OmniSyn |
| | MSE ↓ | | | | | | | | | | | | | | | | | | | | | | | |
| T1→T1c | 0.0160 | 0.0135 | 0.0340 | 0.0135 | 0.0115 | 0.0114 | 0.0112 | **0.0099** | 0.0178 | 0.0149 | 0.0365 | 0.0128 | 0.0126 | 0.0120 | 0.0118 | **0.0116** | 0.0186 | 0.0156 | 0.0380 | 0.0129 | 0.0132 | 0.0126 | 0.0124 | **0.0122** |
| T1→T2 | 0.0148 | 0.0125 | 0.0325 | 0.0115 | 0.0109 | 0.0100 | 0.0095 | **0.0082** | 0.0163 | 0.0138 | 0.0345 | 0.0120 | 0.0119 | 0.0105 | 0.0098 | **0.0093** | 0.0171 | 0.0145 | 0.0362 | 0.0115 | 0.0124 | 0.0111 | 0.0107 | **0.0098** |
| T1c→T1 | 0.0225 | 0.0190 | 0.0360 | 0.0185 | 0.0178 | 0.0175 | 0.0180 | **0.0172** | 0.0248 | 0.0209 | 0.0395 | 0.0208 | 0.0196 | 0.0192 | 0.0194 | **0.0189** | 0.0263 | 0.0222 | 0.0415 | 0.0201 | 0.0206 | 0.0204 | 0.0199 | **0.0198** |
| T1c→T2 | 0.0138 | 0.0118 | 0.0308 | 0.0124 | 0.0098 | 0.0095 | 0.0094 | **0.0080** | 0.0151 | 0.0130 | 0.0339 | 0.0112 | 0.0108 | 0.0101 | 0.0096 | **0.0094** | 0.0160 | 0.0136 | 0.0356 | 0.0108 | 0.0113 | 0.0105 | 0.0103 | **0.0099** |
| T2→T1 | 0.0170 | 0.0142 | 0.0332 | 0.0139 | 0.0118 | 0.0120 | 0.0116 | **0.0105** | 0.0188 | 0.0156 | 0.0358 | 0.0145 | 0.0130 | 0.0123 | 0.0128 | **0.0116** | 0.0196 | 0.0164 | 0.0376 | 0.0129 | 0.0137 | 0.0134 | 0.0130 | **0.0122** |
| T2→T1c | 0.0140 | 0.0119 | 0.0310 | 0.0124 | 0.0096 | 0.0101 | 0.0095 | **0.0088** | 0.0154 | 0.0131 | 0.0345 | 0.0115 | 0.0106 | 0.0100 | 0.0101 | **0.0096** | 0.0161 | 0.0137 | 0.0363 | 0.0109 | 0.0111 | 0.0105 | 0.0103 | **0.0101** |
| | PSNR ↑ | | | | | | | | | | | | | | | | | | | | | | | |
| T1→T1c | 24.50 | 25.80 | 20.00 | 26.81 | 26.40 | 27.23 | 27.64 | **28.05** | 24.00 | 24.80 | 19.40 | 26.08 | 26.10 | 27.25 | 27.43 | **27.60** | 23.60 | 24.40 | 19.10 | 26.68 | 25.60 | 26.34 | 27.01 | **27.10** |
| T1→T2 | 25.00 | 26.20 | 20.40 | 27.20 | 26.80 | 27.60 | 28.00 | **28.40** | 24.60 | 25.50 | 19.80 | 26.46 | 27.30 | 27.63 | 27.79 | **27.95** | 24.10 | 25.10 | 19.50 | 27.06 | 26.00 | 27.19 | 26.83 | **27.40** |
| T1c→T1 | 24.00 | 25.30 | 19.80 | 26.11 | 25.60 | 26.63 | 27.14 | **27.65** | 23.40 | 24.70 | 19.10 | 25.88 | 25.80 | 26.95 | 27.03 | **27.10** | 23.00 | 24.20 | 18.90 | 26.45 | 25.40 | 26.04 | 26.65 | **26.70** |
| T1c→T2 | 25.20 | 26.30 | 20.80 | 26.58 | 26.00 | 27.16 | 27.73 | **28.31** | 24.90 | 25.70 | 20.10 | 26.37 | 27.20 | 27.54 | 27.71 | **27.88** | 24.40 | 25.30 | 19.90 | 27.01 | 25.90 | 27.24 | 26.97 | **27.60** |
| T2→T1 | 24.60 | 25.90 | 20.20 | 25.96 | 25.50 | 26.41 | 26.87 | **27.32** | 24.10 | 25.30 | 19.60 | 25.65 | 26.60 | 26.70 | 26.75 | **26.80** | 23.50 | 24.80 | 19.30 | 26.22 | 25.20 | 25.75 | 26.28 | **26.30** |
| T2→T1c | 25.10 | 26.40 | 20.60 | 26.46 | 25.90 | 27.02 | 27.58 | **28.14** | 24.80 | 25.90 | 20.00 | 26.26 | 27.10 | 27.41 | 27.57 | **27.72** | 24.20 | 25.40 | 19.70 | 26.86 | 25.80 | 27.00 | 26.64 | **27.22** |
| | SSIM ↑ | | | | | | | | | | | | | | | | | | | | | | | |
| T1→T1c | 0.8600 | 0.8700 | 0.7600 | 0.8875 | 0.8870 | 0.8872 | 0.8979 | **0.9080** | 0.8450 | 0.8650 | 0.7400 | 0.8742 | 0.8650 | 0.8648 | 0.8736 | **0.8930** | 0.8200 | 0.8300 | 0.7300 | 0.8564 | 0.8100 | 0.8392 | 0.8306 | **0.8880** |
| T1→T2 | 0.8650 | 0.8720 | 0.7600 | 0.8908 | 0.8980 | 0.8999 | 0.9015 | **0.9120** | 0.8500 | 0.8770 | 0.7400 | 0.8766 | 0.8660 | 0.8662 | 0.8768 | **0.8970** | 0.8260 | 0.8320 | 0.7300 | 0.8458 | 0.8210 | 0.8627 | 0.8713 | **0.8920** |
| T1c→T1 | 0.8400 | 0.8600 | 0.7500 | 0.8782 | 0.8760 | 0.8788 | 0.8873 | **0.8900** | 0.8250 | 0.8550 | 0.7300 | 0.8635 | 0.8620 | 0.8627 | 0.8539 | **0.8740** | 0.8000 | 0.8200 | 0.7200 | 0.8582 | 0.8470 | 0.8509 | 0.8611 | **0.8690** |
| T1c→T2 | 0.8680 | 0.8730 | 0.7600 | 0.8824 | 0.8890 | 0.8830 | 0.8908 | **0.9150** | 0.8470 | 0.8700 | 0.7400 | 0.8788 | 0.8870 | 0.8793 | 0.8881 | **0.9000** | 0.8230 | 0.8470 | 0.7300 | 0.8695 | 0.8500 | 0.8820 | 0.8759 | **0.8920** |
| T2→T1 | 0.8470 | 0.8650 | 0.7600 | 0.8651 | 0.8720 | 0.8765 | 0.8849 | **0.8980** | 0.8280 | 0.8430 | 0.7400 | 0.8414 | 0.8550 | 0.8629 | 0.8602 | **0.8780** | 0.8030 | 0.8200 | 0.7300 | 0.8521 | 0.8320 | 0.8468 | 0.8560 | **0.8730** |
| T2→T1c | 0.8720 | 0.8750 | 0.7700 | 0.8956 | 0.8830 | 0.8943 | 0.9062 | **0.9170** | 0.8480 | 0.8630 | 0.7500 | 0.8896 | 0.8700 | 0.8891 | 0.8798 | **0.8990** | 0.8220 | 0.8470 | 0.7400 | 0.8716 | 0.8500 | 0.8803 | 0.8662 | **0.8920** |

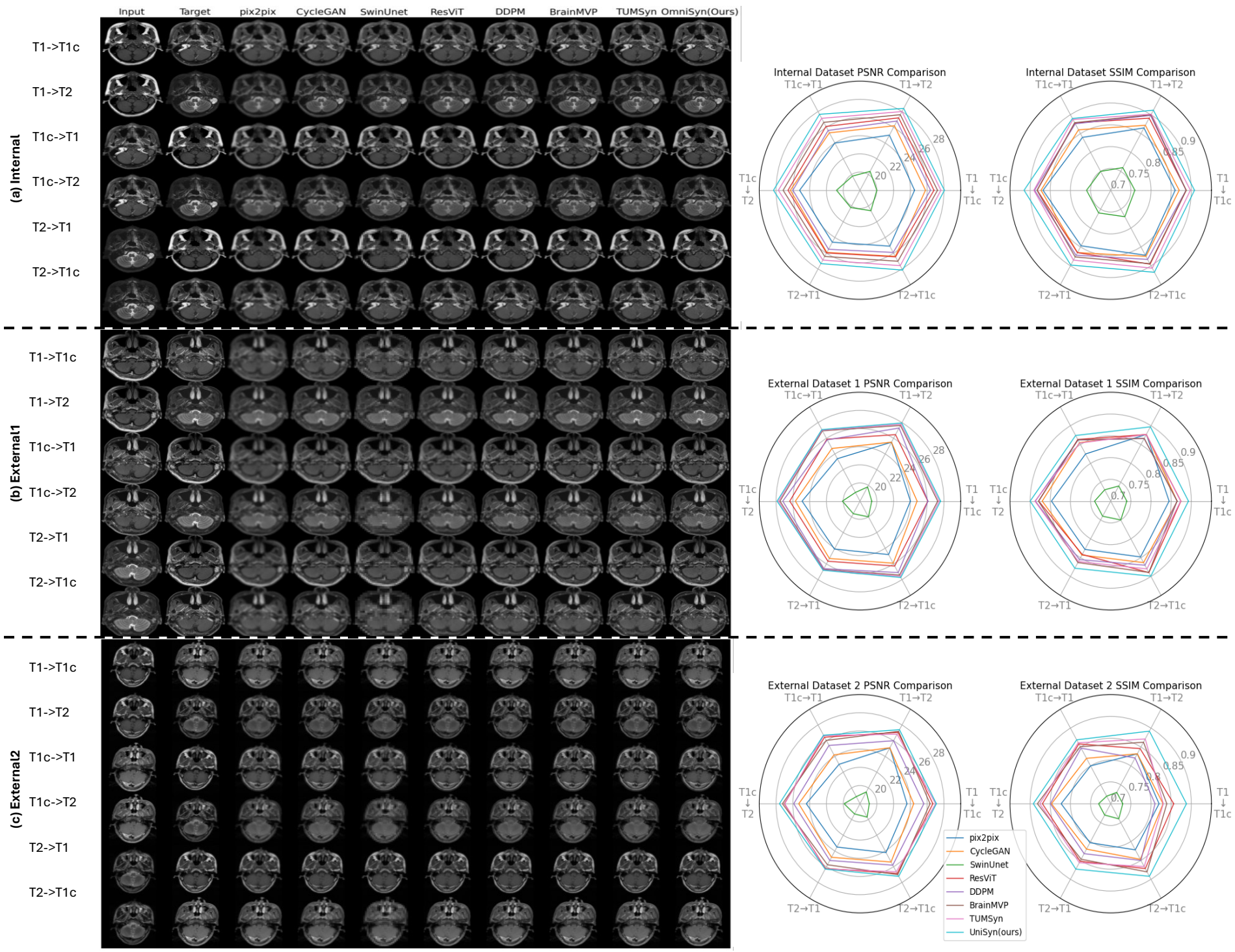

**Figure 2.** Comparison of mainstream image synthesis methods (Pix2pix, CycleGAN, SwinUnet, ResViT, DDPM, BrainMVP, TUMSyn and our OmniSyn) on internal and two external datasets.

**Architectural baseline comparison of generative networks**

In this experiment, we aim to **isolate the effect of network architecture** by comparing different generative backbones under a unified training setting, **without introducing adversarial or contrastive learning strategies**, thereby establishing a fair architectural baseline. The evaluation was performed on the internal dataset and two external datasets (External-1 and External-2) to assess the generalization ability across different data distributions and acquisition settings. All networks were trained under identical configurations, using the same training and validation splits, and evaluated on resampled 3D images with a voxel resolution of $1 \times 1 \times 1$ mm$^3$, resized to $224 \times 224$ per slice. There are seven representative network architectures that have been widely adopted in medical image synthesis and segmentation tasks: (1) UNetr [47], (2) SwinUNet [45], (3) TransUNet [48], (4) ResViT [24], (5) UNet [46], (6) SwinUNetr [49] and (7) SwinUNetrv2 [50]. UNet is a classical convolutional encoder–decoder architecture with symmetric skip connections, serving as a strong baseline for medical image translation. UNetr replaces the convolutional encoder with a Vision Transformer (ViT), while retaining a CNN-based decoder and residual skip connections to capture both global and local representations. SwinUNet employs hierarchical Swin Transformer blocks for both the encoder and decoder, enabling local self-attention within shifted windows. SwinUNetr integrates a Swin Transformer encoder with a CNN decoder, balancing representation power and computational efficiency. SwinUNetrv2 further enhances the original SwinUNetr design by introducing residual convolutional stems and improved feature normalization, achieving better convergence and generalization. TransUNet combines a CNN encoder with a ViT bottleneck and a CNN decoder to model long-range dependencies while maintaining spatial precision. ResViT adopts a residual CNN backbone with modified Transformer blocks to effectively fuse spatial and contextual features. All competing models were trained under identical experimental settings and data partitions to ensure a fair comparison with our proposed framework. As illustrated in **Figure 3**, ResViT exhibits the poorest visual performance, with the generated images largely failing to preserve meaningful anatomical structures, resulting in blurred and anatomically implausible appearances. UNetr, SwinUNet, and TransUNet produce comparatively inferior results, where the overall anatomical layout can be partially recognized, but fine-grained textures remain indistinct and boundaries are poorly defined. In contrast, U-Net, SwinUNetr, and SwinUNetrV2 achieve the best visual quality, generating images with more coherent anatomical structures and improved texture fidelity; however, despite their superior performance, certain subtle anatomical details and high-frequency textures are still insufficiently resolved, indicating room for further improvement.

**Integration of GAN strategy into backbone networks**

Building upon the architectural baseline results, this experiment investigates whether **adversarial learning can further improve synthesis realism** by integrating a ResNet-based discriminator into the training of the three top-performing backbones: UNet, SwinUNetr, and SwinUNetrv2 (**Figure 3**, last six columns). Across all experimental settings, adding the GAN strategy **consistently improves performance**. The largest gains are observed in SSIM and perceptual quality assessments, indicating that adversarial supervision helps recover subtle tissue boundaries and reduces over-smoothing commonly seen in purely regression-based training. For example, UNet-GAN shows enhanced reconstruction of cortical folds, while SwinUNetr-GAN more accurately synthesizes small vessels and deep-brain structures. The improvement is most evident for SwinUNetrv2, whose hierarchical transformer encoder already captures global context effectively; the adversarial loss further helps refine local details. Importantly, the GAN augmentation does not introduce instability or hallucinated features, likely due to the moderate discriminator capacity and carefully balanced training schedule.

**Table 2. Comparison of seven different baseline networks and adding GAN effects.**

|  | Unetr | Swinunet | Transunet | ResViT | Unet | Unet+GAN | Swinunetr | Swinunetr+GAN | Swinunetrv2 | Swinunetrv2+GAN |
|---|---|---|---|---|---|---|---|---|---|---|
| MSE ↓ | | | | | | | | | | |
| T1->T1c | 0.0143 | 0.0135 | 0.0158 | 0.0261 | 0.0125 | 0.0097 | 0.0134 | **0.0094** | 0.0138 | 0.0095 |
| T1->T2 | 0.0124 | 0.0120 | 0.0138 | 0.0318 | 0.0094 | **0.0091** | 0.0108 | 0.0095 | 0.0106 | 0.0093 |
| T1c->T1 | 0.0245 | 0.0239 | 0.0274 | 0.0450 | 0.0196 | **0.0153** | 0.0214 | 0.0168 | 0.0217 | 0.0157 |
| T1c->T2 | 0.0116 | 0.0117 | 0.0145 | 0.0318 | 0.0093 | **0.0084** | 0.0101 | 0.0094 | 0.0105 | 0.0094 |
| T2->T1 | 0.0253 | 0.0257 | 0.0314 | 0.0452 | 0.0178 | **0.0125** | 0.0212 | 0.0156 | 0.0221 | 0.0175 |
| T2->T1c | 0.0124 | 0.0120 | 0.0160 | 0.0245 | 0.0105 | **0.0096** | 0.0115 | 0.0103 | 0.0116 | 0.0101 |
| PSNR ↑ | | | | | | | | | | |
| T1->T1c | 18.82 | 19.06 | 18.33 | 16.04 | 19.52 | **20.71** | 19.15 | 20.12 | 19.01 | 21.21 |
| T1->T2 | 19.66 | 19.85 | 19.16 | 15.14 | 20.99 | **22.66** | 20.43 | 21.56 | 20.45 | 22.34 |
| T1c->T1 | 16.38 | 16.46 | 15.97 | 13.55 | 17.43 | 18.56 | 17.05 | 18.11 | 16.97 | **19.04** |
| T1c->T2 | 20.11 | 19.99 | 19.12 | 15.13 | 21.16 | **22.12** | 20.80 | 21.95 | 20.48 | 21.77 |
| T2->T1 | 16.27 | 16.16 | 15.60 | 13.56 | 17.83 | **18.68** | 17.02 | 18.66 | 16.82 | 18.59 |
| T2->T1c | 19.48 | 19.58 | 18.35 | 16.36 | 20.32 | **21.46** | 19.85 | 20.86 | 19.79 | 20.99 |
| SSIM ↑ | | | | | | | | | | |
| T1->T1c | 0.5079 | 0.4946 | 0.4491 | 0.3609 | 0.5207 | **0.5712** | 0.5042 | 0.5448 | 0.4975 | 0.5647 |
| T1->T2 | 0.5046 | 0.4870 | 0.4748 | 0.3309 | 0.5479 | **0.5946** | 0.5222 | 0.5678 | 0.5133 | 0.5714 |
| T1c->T1 | 0.2960 | 0.3789 | 0.3061 | 0.2207 | 0.4909 | **0.5349** | 0.3500 | 0.4150 | 0.3820 | 0.4221 |
| T1c->T2 | 0.5589 | 0.5374 | 0.5203 | 0.3303 | 0.5855 | **0.6135** | 0.5515 | 0.5768 | 0.5192 | 0.5777 |
| T2->T1 | 0.3060 | 0.4103 | 0.3019 | 0.2268 | 0.4806 | **0.5253** | 0.3774 | 0.4320 | 0.4011 | 0.4657 |
| T2->T1c | 0.5491 | 0.5454 | 0.4200 | 0.4183 | 0.5760 | **0.6173** | 0.5425 | 0.5963 | 0.5393 | 0.5936 |
| Parameters | 25.55M | 27.17M | 67.01M | 123.35M | **14.78M** | +21.8M | 25.14M | +21.8M | 28.66M | +21.8M |

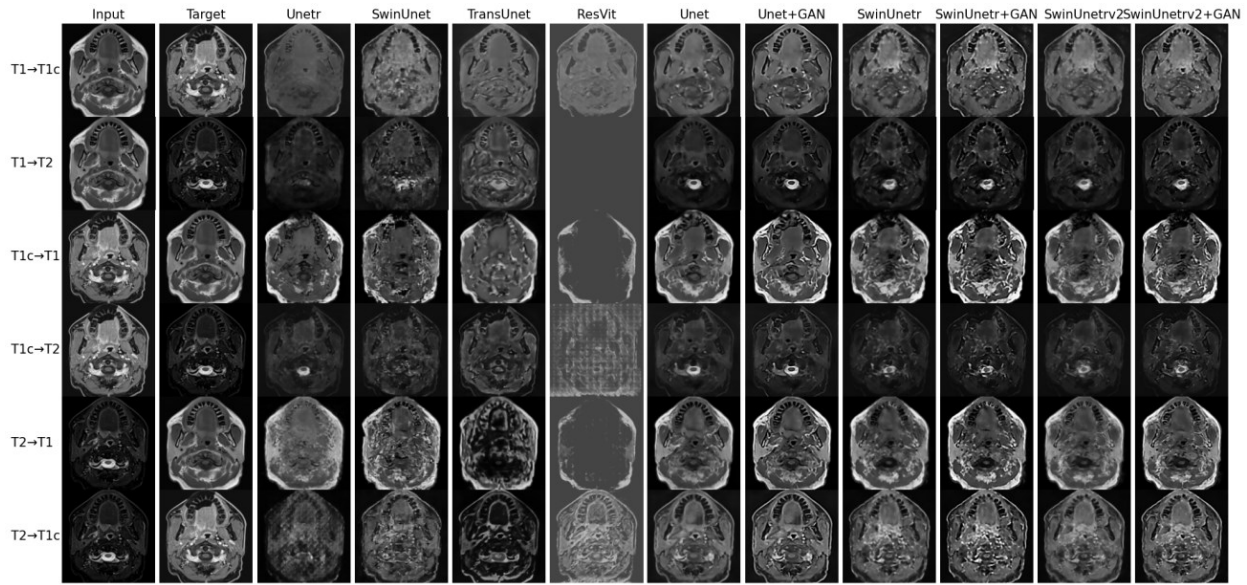

**Figure 3.** Comparison of only seven different baseline networks (Unetr, SwinUnet, TransUnet, ResViT, Unet, SwinUnetr and SwinUnetrv2). Last 6 columns are results without and with GAN strategy.

Overall, across the three comparison settings, each experiment plays a distinct role in shaping our framework. The mainstream comparison summarizes how all representative synthesis methods perform in their final forms. The baseline comparison helps us determine the three strongest architectural foundations. The GAN-based comparison demonstrates that adversarial supervision provides meaningful improvements, motivating our later introduction of language-based supervision as a more stable semantic discriminator.

**Ablation study**

**Comparison of different architectures**

We first evaluate the influence of architectural unification on multi-modality MRI synthesis by comparing three progressively integrated designs. The **initial configuration** (three encoders three decoders, **3E3D**) employs separate encoders and decoders for every source–target modality pair, effectively treating each mapping as an independent task. Although this fully decoupled framework provides maximal capacity for modality-specific information, it results in redundant parameters and limited generalization. The lack of shared anatomical representation leads to inconsistent performance across datasets and hampers robustness when faced with unfamiliar contrasts or external domain shifts. To address these limitations, the **second configuration** (one encoder three decoders, **1E3D**) introduces a unified encoder while retaining separate decoders for each target modality. This design significantly improves synthesis fidelity, demonstrating that a shared encoder can capture modality-invariant anatomical structures while still allowing decoders to specialize in contrast-specific mappings. However, the need for independent decoders still restricts flexibility and prevents the model from leveraging correlations among target modalities. The **final**

**configuration**, fully adopted by OmniSyn (one encoder one decoder and one condition, **1E1D1Cond**), unifies both the encoder and decoder into a single backbone conditioned on a modality label. This architecture learns a generalizable representation space capable of supporting all cross-modality conversions within a single model. Notably, performance remains comparable to the previous two designs across internal and external datasets. At the same time, this unified model reduces the number of parameters, streamlines the training workflow, and substantially enhances generalization and scalability. These results show that a fully unified architecture is sufficient to support high-fidelity MRI synthesis while offering clear advantages in efficiency and extensibility. Configurations and results are shown in **Figure 4**.

**Table 3. Comparison of 3 architectures.**

|  | Unet (3E3D) | (1E3D) | (1E1D1Cond) | swinunetr(3E3D) | (1E3D) | (1E1D1Cond) | swinunetrv2(3E3D) | (1E3D) | (1E1D1Cond) |
|---|---|---|---|---|---|---|---|---|---|
| | | | | MSE ↓ | | | | | |
| T1->T1c | 0.0125 | 0.0124 | 0.0126 | 0.0134 | 0.0133 | 0.0134 | 0.0138 | 0.0136 | 0.0133 |
| T1->T2 | 0.0094 | 0.0095 | 0.0093 | 0.0108 | 0.0111 | 0.0112 | 0.0106 | 0.0104 | 0.0104 |
| T1c->T1 | 0.0196 | 0.0196 | 0.0195 | 0.0214 | 0.0215 | 0.0216 | 0.0217 | 0.0215 | 0.0210 |
| T1c->T2 | 0.0093 | 0.0093 | 0.0095 | 0.0101 | 0.0106 | 0.0107 | 0.0105 | 0.0106 | 0.0103 |
| T2->T1 | 0.0178 | 0.0176 | 0.0174 | 0.0212 | 0.0227 | 0.0238 | 0.0221 | 0.0221 | 0.0222 |
| T2->T1c | 0.0105 | 0.0104 | 0.0105 | 0.0115 | 0.0116 | 0.0117 | 0.0116 | 0.0111 | 0.0110 |
| | | | | PSNR ↑ | | | | | |
| T1->T1c | 19.52 | 19.52 | 19.53 | 19.15 | 19.15 | 19.17 | 19.01 | 18.95 | 19.17 |
| T1->T2 | 20.99 | 21.00 | 21.04 | 20.43 | 20.16 | 20.19 | 20.45 | 21.04 | 20.53 |
| T1c->T1 | 17.43 | 17.46 | 17.42 | 17.05 | 17.03 | 16.99 | 16.97 | 16.88 | 17.09 |
| T1c->T2 | 21.16 | 21.09 | 21.04 | 20.80 | 20.88 | 20.44 | 20.48 | 20.54 | 20.51 |
| T2->T1 | 17.83 | 18.01 | 17.87 | 17.02 | 16.98 | 16.56 | 16.82 | 16.78 | 16.86 |
| T2->T1c | 20.32 | 20.22 | 20.28 | 19.85 | 19.75 | 19.69 | 19.79 | 19.89 | 20.00 |
| | | | | SSIM ↑ | | | | | |
| T1->T1c | 0.5207 | 0.5268 | 0.5322 | 0.5042 | 0.5044 | 0.5046 | 0.4975 | 0.5046 | 0.5062 |
| T1->T2 | 0.5479 | 0.5446 | 0.5440 | 0.5222 | 0.5050 | 0.5055 | 0.5133 | 0.5195 | 0.5173 |
| T1c->T1 | 0.4909 | 0.3569 | 0.3683 | 0.3500 | 0.3654 | 0.3993 | 0.3820 | 0.3939 | 0.4191 |
| T1c->T2 | 0.5855 | 0.5789 | 0.5867 | 0.5515 | 0.5464 | 0.5373 | 0.5192 | 0.5047 | 0.5021 |
| T2->T1 | 0.4806 | 0.5036 | 0.5444 | 0.3774 | 0.4056 | 0.4134 | 0.4011 | 0.4281 | 0.4343 |
| T2->T1c | 0.5760 | 0.5776 | 0.5751 | 0.5425 | 0.5312 | 0.5208 | 0.5393 | 0.5406 | 0.5453 |

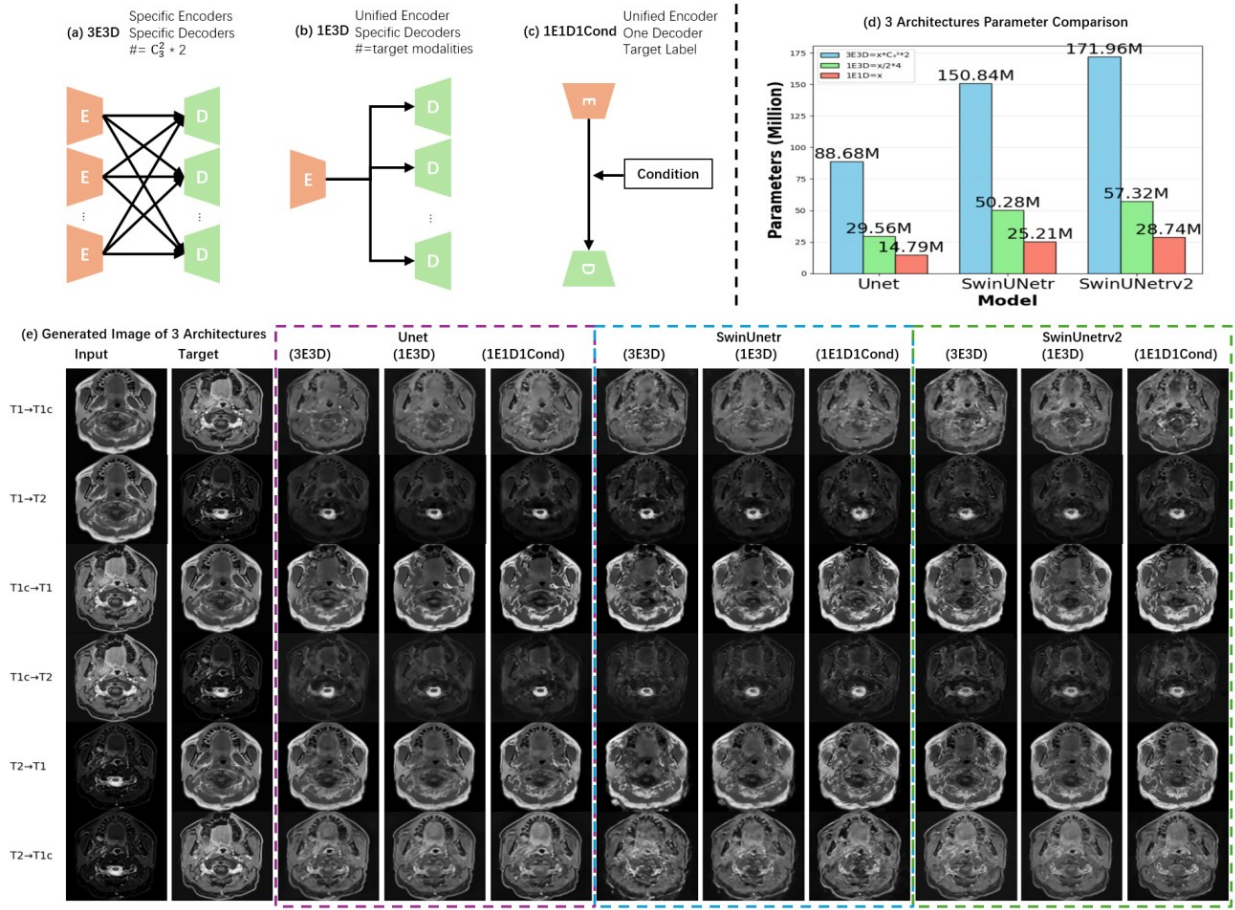

**Figure 4.** Three architectures' configuration and generated visualization results. (a)-(c) are three illustrations of different architecture combinations. (d) shows the parameters' changes in different configurations. (e) shows the generated images of this ablation study. Purple, blue, and green boxes indicate results from U-Net, SwinUNETR, and SwinUNETRv2, respectively. Each has three columns representing different configurations.

**Effects of different training strategies**

Beyond architectural design, we also assess the contributions of our two proposed training strategies: contrastive pretraining and CLIP-guided vision–language supervision. When **neither strategy is used**, the model relies exclusively on pixel-level regression. Although reasonable performance is achieved, the resulting representations remain modality-dependent and lack strong structural consistency, leading to reduced robustness when synthesizing slices with varying anatomical complexity or when evaluated on external datasets. **Adding contrastive pretraining** markedly enhances the encoder's ability to learn slice-consistent and modality-invariant features. By pulling together representations of the same anatomical slice across different modalities, the model develops a structurally aligned feature space that stabilizes the synthesis process. However, contrastive learning alone does not provide semantic guidance on how modality-specific appearance should differ. **Incorporating CLIP-based supervision** addresses this gap by introducing a high-level constraint that associates each synthesized output with its corresponding textual modality description. This acts as a discriminator-like semantic signal, improving perceptual realism and ensuring that synthesized contrasts follow the correct modality characteristics. When **both strategies are combined**, their complementary strengths yield the most substantial improvements in quantitative accuracy, perceptual quality, and cross-dataset generalization. This synergy highlights the importance of coupling structural alignment with semantic supervision in building a unified and reliable foundation model for MRI synthesis. See **Figure 5** and **Table 3** (a, b and c mean internal, external1 and external2 datasets).

**Table 4.** Effects of two training strategies – Unified Encoder Contrastive Learning and CLIP supervised Training.

| | Internal | | | | External 1 | | | | External 2 | | | |
|---|---|---|---|---|---|---|---|---|---|---|---|---|
| | Baseline | Stategy1 | Stategy2 | 1+2 | Baseline | Stategy1 | Stategy2 | 1+2 | Baseline | Stategy1 | Stategy2 | 1+2 |
| MSE ↓ | | | | | | | | | | | | |
| T1→T1c | 0.0138 | 0.0115 | 0.0108 | **0.0099** | 0.0218 | 0.0203 | 0.0191 | **0.0180** | 0.0224 | 0.0210 | 0.0198 | **0.0187** |
| T1→T2 | 0.0106 | 0.0098 | 0.0090 | **0.0082** | 0.0206 | 0.0194 | 0.0183 | **0.0172** | 0.0214 | 0.0201 | 0.0189 | **0.0178** |
| T1c→T1 | 0.0217 | 0.0197 | 0.0184 | **0.0172** | 0.0321 | 0.0300 | 0.0287 | **0.0273** | 0.0318 | 0.0303 | 0.0290 | **0.0279** |
| T1c→T2 | 0.0105 | 0.0096 | 0.0088 | **0.0080** | 0.0215 | 0.0202 | 0.0189 | **0.0178** | 0.0204 | 0.0191 | 0.0180 | **0.0171** |
| T2→T1 | 0.0221 | 0.0124 | 0.0113 | **0.0105** | 0.0291 | 0.0274 | 0.0258 | **0.0246** | 0.0312 | 0.0295 | 0.0279 | **0.0266** |
| T2→T1c | 0.0116 | 0.0105 | 0.0096 | **0.0088** | 0.0226 | 0.0212 | 0.0200 | **0.0190** | 0.0231 | 0.0218 | 0.0207 | **0.0197** |
| PSNR ↑ | | | | | | | | | | | | |
| T1→T1c | 19.01 | 23.52 | 26.10 | **28.05** | 18.11 | 22.40 | 25.40 | **27.60** | 18.66 | 22.90 | 25.80 | **28.10** |
| T1→T2 | 20.45 | 24.10 | 26.55 | **28.40** | 19.46 | 23.00 | 25.80 | **27.95** | 19.29 | 23.50 | 26.20 | **28.45** |
| T1c→T1 | 16.97 | 21.05 | 25.30 | **27.65** | 15.77 | 19.90 | 24.50 | **27.10** | 14.97 | 19.10 | 24.00 | **27.00** |
| T1c→T2 | 20.48 | 24.22 | 26.70 | **28.31** | 19.32 | 23.05 | 25.83 | **27.88** | 18.98 | 23.30 | 26.00 | **28.32** |
| T2→T1 | 16.82 | 20.70 | 24.90 | **27.32** | 15.24 | 19.50 | 24.15 | **26.80** | 16.32 | 20.55 | 25.10 | **27.60** |
| T2→T1c | 19.79 | 23.80 | 26.20 | **28.14** | 18.29 | 22.55 | 25.60 | **27.72** | 18.96 | 23.20 | 25.95 | **28.22** |
| SSIM ↑ | | | | | | | | | | | | |
| T1→T1c | 0.4975 | 0.7020 | 0.8450 | **0.9080** | 0.4725 | 0.6900 | 0.8400 | **0.9030** | 0.4551 | 0.6840 | 0.8380 | **0.9010** |
| T1→T2 | 0.5133 | 0.7150 | 0.8560 | **0.9120** | 0.4989 | 0.7080 | 0.8520 | **0.9100** | 0.5088 | 0.7200 | 0.8620 | **0.9130** |
| T1c→T1 | 0.3820 | 0.5980 | 0.8120 | **0.8900** | 0.3511 | 0.5800 | 0.7980 | **0.8840** | 0.3636 | 0.5900 | 0.8060 | **0.8860** |
| T1c→T2 | 0.5192 | 0.7200 | 0.8670 | **0.9150** | 0.4962 | 0.7030 | 0.8580 | **0.9090** | 0.5029 | 0.7100 | 0.8580 | **0.9080** |
| T2→T1 | 0.4011 | 0.6200 | 0.8280 | **0.8980** | 0.3964 | 0.6120 | 0.8180 | **0.8930** | 0.3890 | 0.6150 | 0.8200 | **0.8950** |
| T2→T1c | 0.5393 | 0.7360 | 0.8720 | **0.9170** | 0.5012 | 0.7200 | 0.8650 | **0.9140** | 0.5248 | 0.7350 | 0.8720 | **0.9160** |

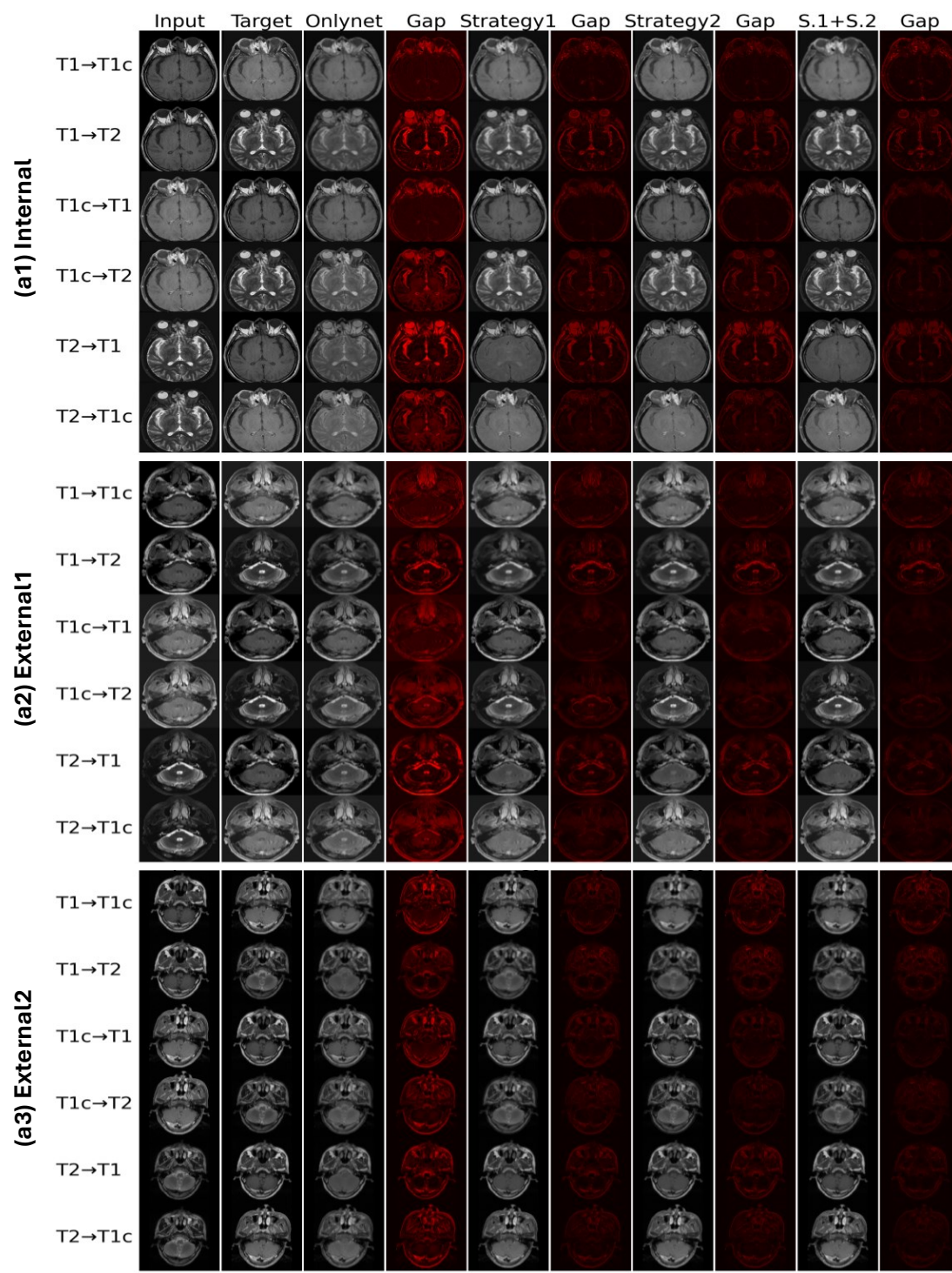
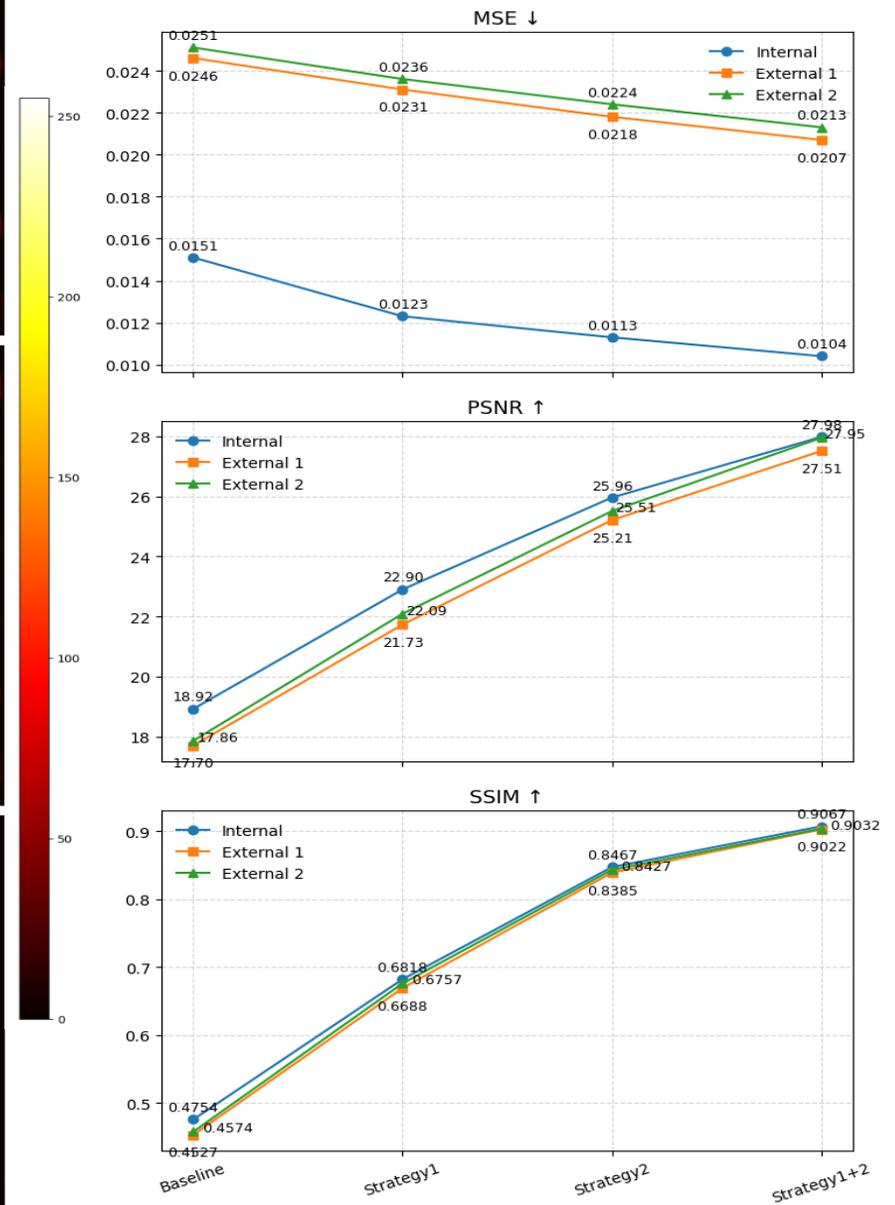

**Figure 5.** Ablation study of two training strategies on three datasets. Left panels (a1)–(a3) present visual results under four settings: without either strategy, with strategy 1, with strategy 2, and with both strategies applied. The gap columns illustrate the differences between predictions and ground truth. Right panel (b) shows the trends of three quantitative metrics.

**Intermediate feature analysis of two training strategies**

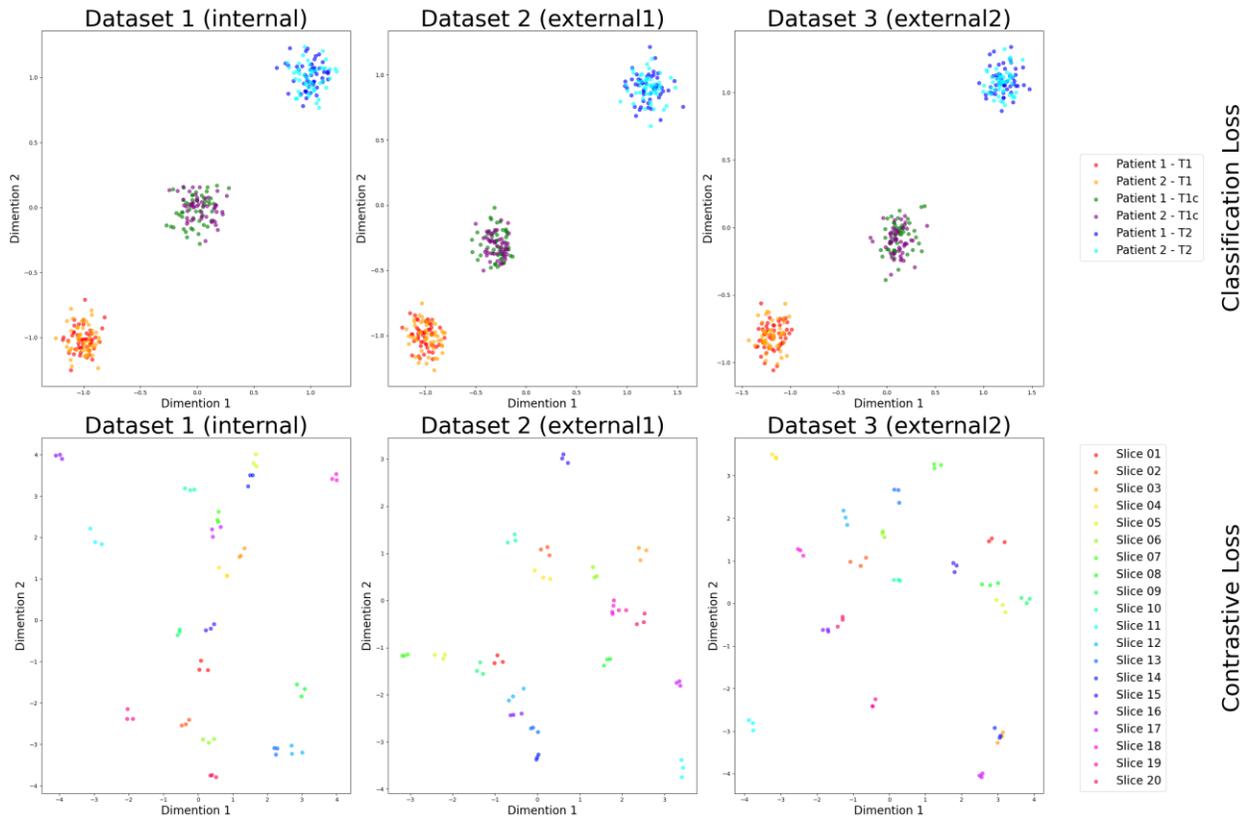

**Figure 6.** Feature cluster results using classification conditions and contrastive learning convergence conditions. The first line means the same modality images are clustered together and different ones are separated. The second line means the same slice different modality images are clustered together.

To better understand the behavior of the unified encoder and verify the effectiveness of contrastive pretraining, we perform an intermediate analysis by visualizing the latent representations produced by the encoder. Using PCA-based dimensionality reduction [51] and clustering-oriented embedding visualization (**Figure 6**), we project the output vectors—generated by the two encoders after classification training and contrastive pretraining—onto a two-dimensional space for slices from different modalities. The resulting clusters reveal a clear and consistent pattern: images from the same anatomical slice but originating from different modalities are embedded close to one another, whereas images from different slices—even within the same modality—tend to separate into distinct clusters. This is benefit for our synthesis mission, which is different from the classification mission.

This behavior confirms that the contrastive objective successfully drives the encoder to learn *slice-consistent* and *modality-invariant representations*. Rather than focusing on low-level appearance differences across modalities, the encoder learns to emphasize anatomical structure and semantic content. Such alignment is crucial for cross-modality synthesis, where the model must understand that T1, T2, and other MRI sequences represent different manifestations of the same underlying anatomy. These visualization findings (**Figure 6** second line) validate the design of our pretraining strategy: contrastive learning equips the encoder with a strong, geometry-aware feature space that facilitates downstream synthesis and stabilizes training across diverse datasets. This structure-preserving representation space also contributes to OmniSyn's robustness when handling out-of-distribution (OOD) inputs.

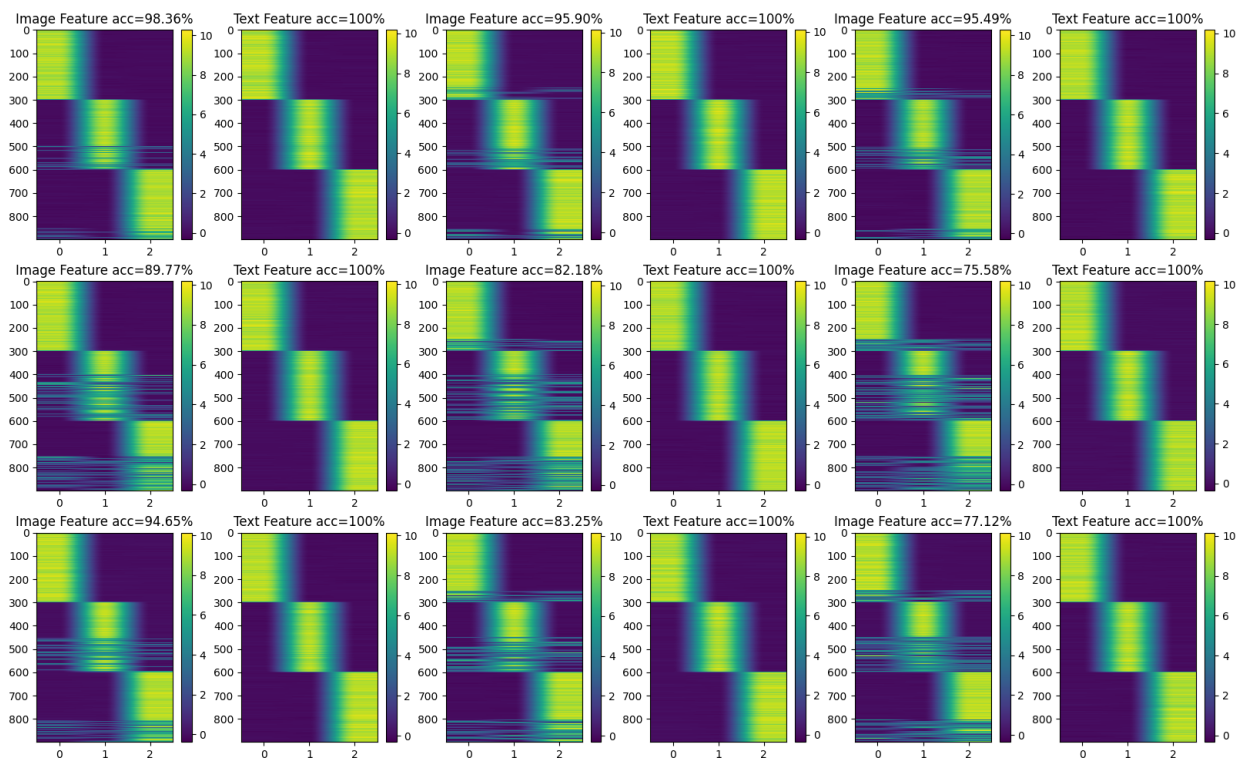

**Figure 7.** Visualization of three CLIPs classification accuracy for text and images on internal & two external datasets, after finetuning. The vertical axis represents sample indices, while the horizontal axis denotes modal categories 012 corresponding to T1, T1c, and T2. The brighter the spectral region, the higher the probability of a particular modal category.

To further investigate the role of language–vision alignment, we compare three pretrained vision–language models—**CLIP** [33], **MedCLIP** [34], and **RadCLIP** [35]—under both zero-shot and fine-tuned settings. When evaluated in a strict zero-shot manner, all models initially exhibit **near-zero accuracy** in predicting modality labels from either images or text prompts. This failure is primarily due to **label mismatches** between our dataset and the terminology used during the original pretraining of these models. For example, our modality names (T1, T1c, T2), label names (0, 1, 2) and textual descriptions differ from the canonical medical terminology present in MedCLIP or RadCLIP training corpora, making zero-shot inference unreliable.

After modality-aligned fine-tuning on our internal dataset, performance improves dramatically. Text-to-modality prediction reaches 100% accuracy across all three models, demonstrating that the text encoder can be easily adapted to our modality vocabulary. For image-based modality classification, CLIP achieves the highest accuracy, followed by RadCLIP, while MedCLIP shows the lowest performance. On external datasets 1 and 2, accuracy decreases due to domain shift; however, CLIP maintains up to ~90% accuracy, significantly outperforming both MedCLIP and RadCLIP. These results show that (1) fine-tuning is necessary for modality alignment, and (2) CLIP initialization provides the strongest generalization ability across domains. This motivates our use of CLIP-based supervision as a *semantically informed discriminator-like signal* in the OmniSyn framework. Details of quantitative and visualization are shown on **Table 5** and **Figure 7**.

**Table 5.** Quantitative of similarities between feature vectors and classification accuracy using three versions of CLIP.

| Cosine Similarity [0~1] | Random | Classification | Contrastive Learning |
| --- | --- | --- | --- |
| T1<->T1c | 0.6268 | 0.6653 | **0.8879** |
| T1<->T2 | 0.4578 | 0.4312 | **0.8751** |
| T1c<->T2 | 0.4457 | 0.4245 | **0.8765** |
| Acc (%) | CLIP Fine Tuning | MedCLIP Fine Tuning | RadCLIP Fine Tuning |
| Text | 100 | 100 | 100 |
| Image(internal) | 98.36 | 95.49 | 95.90 |
| Image(external1) | 89.77 | 75.58 | 82.18 |
| Image(external2) | 94.65 | 77.84 | 81.25 |

**Robustness evaluation**

To comprehensively assess the robustness of the proposed foundation model, we systematically evaluated its behavior under controlled noise perturbations, multiple corruption types, and clinically relevant domain shifts.

**Robustness to noise and image corruption**

First, we introduced **motion artifacts** [52], **down-sampling**, **Gaussian noise** [53], and **Rician noise** [54] to the input images at three predefined severity levels—**minor**, **moderate**, and **severe**—to simulate real-world acquisition imperfections. Across all corruption types, the model demonstrated a degradation curve: performance decreased steadily with increasing corruption but remained consistently higher than competing baselines. Remarkably, even under severe motion artifacts and heavy noise, the unified encoder–decoder architecture maintained structural fidelity and contrast consistency, indicating that the model had learned modality-invariant and texture-resilient representations. The findings highlight that our unified design not only supports multi-modality synthesis but also naturally benefits from shared robustness priors learned across modalities. Moreover, the model showed strong resilience to down-sampling, preserving anatomical boundaries even when spatial resolution was substantially reduced, confirming that the latent space was robust to degradation of fine details. See on **Table 6** and **Figure 8**.

**Table 6.** Robustness test about 4 types and 3 levels of corruption.

| | Motion Artifact | | Prediction | | Down Sampling | | Prediction | | Gaussian Noise | | Prediction | | Rician Noise | | Prediction | |
|---|---|---|---|---|---|---|---|---|---|---|---|---|---|---|---|---|
| | PSNR | SSIM | PSNR | SSIM | PSNR | SSIM | PSNR | SSIM | PSNR | SSIM | PSNR | SSIM | PSNR | SSIM | PSNR | SSIM |
| **Internal** | | | | | | | | | | | | | | | | |
| **Minor** | 37.73 | 0.9353 | 34.45 | 0.9690 | 28.90 | 0.9349 | 28.37 | 0.8320 | 28.71 | 0.6421 | 30.46 | 0.8754 | 27.83 | 0.6192 | 29.26 | 0.8567 |
| **Moderate** | 26.83 | 0.7357 | 25.64 | 0.7498 | 25.00 | 0.8461 | 25.00 | 0.7426 | 19.89 | 0.3262 | 25.53 | 0.7421 | 18.12 | 0.3072 | 23.77 | 0.6436 |
| **Severe** | 23.12 | 0.6753 | 22.21 | 0.6500 | 22.69 | 0.7550 | 22.43 | 0.6953 | 14.63 | 0.1745 | 23.23 | 0.7057 | 11.62 | 0.1406 | 19.33 | 0.4296 |
| **External 1** | | | | | | | | | | | | | | | | |
| **Minor** | 30.45 | 0.8755 | 28.36 | 0.8446 | 29.67 | 0.9382 | 27.44 | 0.8461 | 28.53 | 0.6813 | 28.77 | 0.8645 | 27.90 | 0.6650 | 28.94 | 0.8674 |
| **Moderate** | 28.72 | 0.8529 | 25.01 | 0.7982 | 25.80 | 0.8520 | 23.82 | 0.7519 | 19.60 | 0.3501 | 25.23 | 0.7694 | 18.26 | 0.3328 | 25.37 | 0.8084 |
| **Severe** | 19.94 | 0.5075 | 20.71 | 0.6612 | 23.45 | 0.7637 | 20.83 | 0.6953 | 14.28 | 0.1814 | 20.62 | 0.6227 | 11.93 | 0.1509 | 19.00 | 0.4878 |
| **External 2** | | | | | | | | | | | | | | | | |
| **Minor** | 29.55 | 0.7716 | 31.43 | 0.9166 | 29.87 | 0.9587 | 35.28 | 0.9524 | 29.72 | 0.5822 | 33.20 | 0.9359 | 27.03 | 0.4825 | 35.02 | 0.9174 |
| **Moderate** | 25.39 | 0.6963 | 30.07 | 0.8797 | 26.11 | 0.9047 | 33.61 | 0.9411 | 20.39 | 0.2727 | 28.65 | 0.7690 | 16.96 | 0.2352 | 26.35 | 0.5756 |
| **Severe** | 20.78 | 0.5036 | 25.96 | 0.7230 | 23.75 | 0.8436 | 27.25 | 0.8357 | 14.68 | 0.1405 | 24.81 | 0.5041 | 10.79 | 0.1077 | 20.12 | 0.3804 |

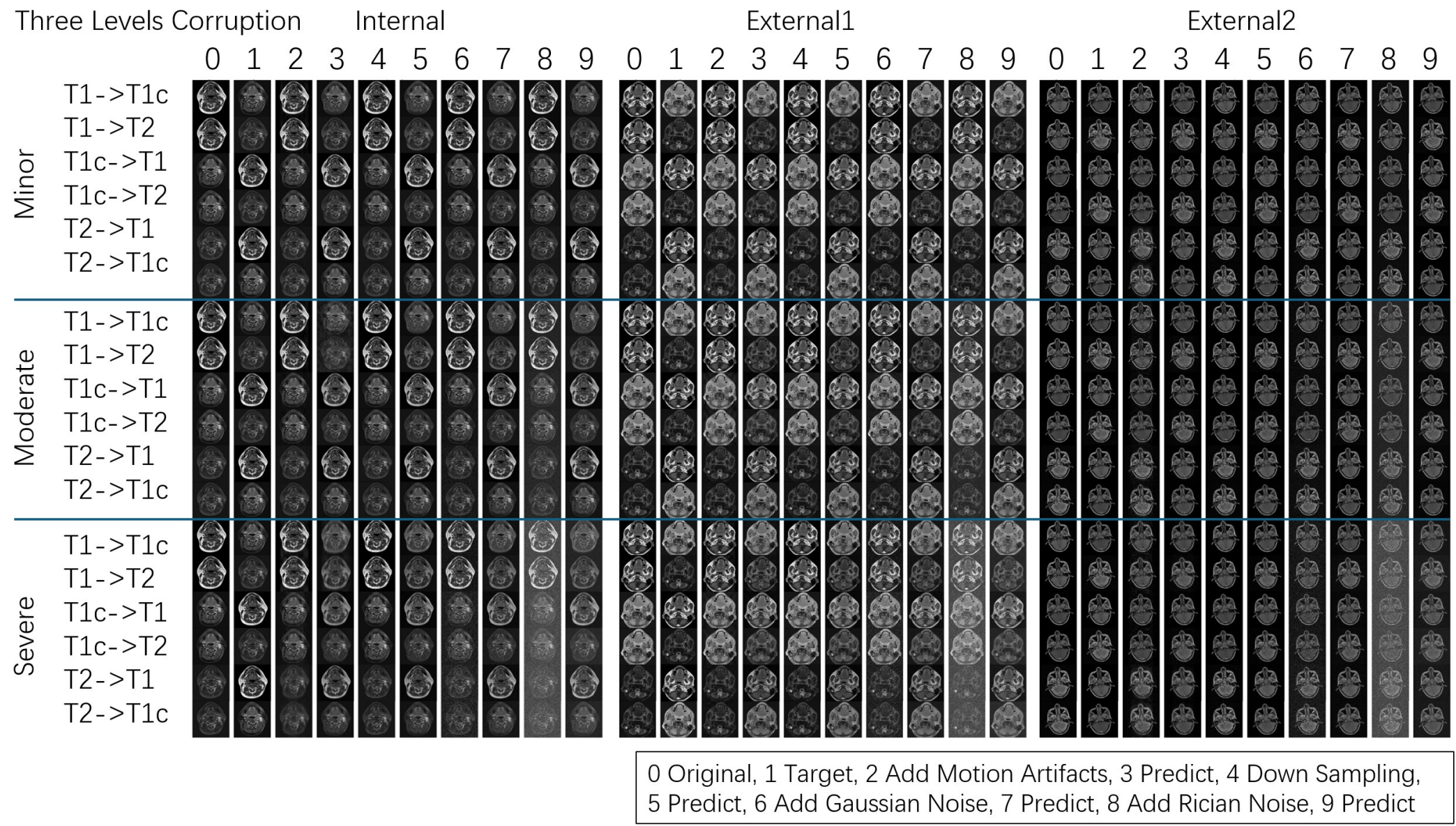

**Figure 8.** Robustness test during three levels of corruption (minor, moderate and severe) on internal, external1 and external2 datasets. We applied motion artifacts, down sampling, Gaussian noise and Rician noise to original images and generated the target modality MRI images.

**Robustness under different domain shifts**

Beyond corruption-based robustness, we quantified generalization under domain shifts by evaluating the model on **two external datasets** with markedly different acquisition environments. **External-1**, collected from **25 hospitals**, represents a heterogeneous multi-institutional dataset with diverse scanners, protocols, and patient populations. **External-2**, originating from a single institution with a unique acquisition workflow, captures a distinct domain not seen during training. The proposed model achieved stable performance across both datasets, with only mild performance drops relative to the internal test set. This indicates strong domain generalization ability, largely attributed to the unified encoder's capacity to capture modality-agnostic structures and the contrastive alignment introduced during training. Compared with baseline architecture, our framework exhibited noticeably greater robustness to cross-site variability, underscoring its suitability for deployment in large-scale, real-world clinical environments. See on the *Overall performance* section.

**Evaluation on downstream tasks**

To further validate the practical utility of the proposed foundation model on NPC, we conducted comprehensive downstream evaluations on two representative segmentation tasks. These tasks were designed to assess whether the synthesized modalities preserve task-relevant anatomical cues and whether the learned representations generalize to clinically meaningful applications.

**Table 7.** Downstream tasks segmentation accuracy. First task is NPC ROI segmentation, and second task is brain tissue segmentation.

| DSC [0~1] | Source / Target | Original | Generated from T1c/T1/T1 | Generated from T2/T2/T1c |
|---|---|---|---|---|
| External 2 Dataset | T1 ROI segmentation | 0.8657 ± 0.2900 | 0.8655 ± 0.2897 | 0.8686 ± 0.2876 |
| | T1c ROI segmentation | 0.8611 ± 0.3008 | 0.8600 ± 0.3019 | 0.8602 ± 0.3015 |
| | T2 ROI segmentation | 0.8604 ± 0.2962 | 0.8642 ± 0.2923 | 0.8615 ± 0.2965 |
| DSC [0~1] | Source / Target | Original | Generated from T1c/T1/T1 | Generated from T2/T2/T1c |
| External 1 Dataset | T1 Tissue segmentation | - | 0.8515 ± 0.0796 | 0.7776 ± 0.0569 |
| | T1c Tissue segmentation | - | 0.8501 ± 0.0629 | 0.7891 ± 0.0405 |
| | T2 Tissue segmentation | - | 0.8534 ± 0.0734 | 0.8425 ± 0.0554 |

**Application to NPC segmentation**

We first evaluated the model on NPC segmentation using the External-2 test dataset [55], which was acquired from a distinct clinical center with imaging characteristics different from the training distribution. The ground truth of NPC was labeled by professional clinicians. Segmentation models trained on real images were directly applied to synthesized images without any fine-tuning, thereby providing a stringent test of synthesis fidelity and clinical usability. The results demonstrate that synthesis-based segmentation achieves performance comparable to, and in some cases exceeding, segmentation on real images—particularly in challenging tumor margins where texture and contrast transitions are subtle. This confirms that the synthesized modalities not only reconstruct global structures but also faithfully preserve tumor-specific features essential for precise delineation. Importantly, as evidenced by **Table 7** and **Figure 9**, the unified encoder–decoder design achieves segmentation performance that is highly comparable to that obtained using real images, demonstrating strong robustness under domain shift. The resulting segmentation outcomes are essentially on par with those derived from real-image inputs, indicating that the unified representations and CLIP-based supervision effectively mitigate performance degradation across domains.

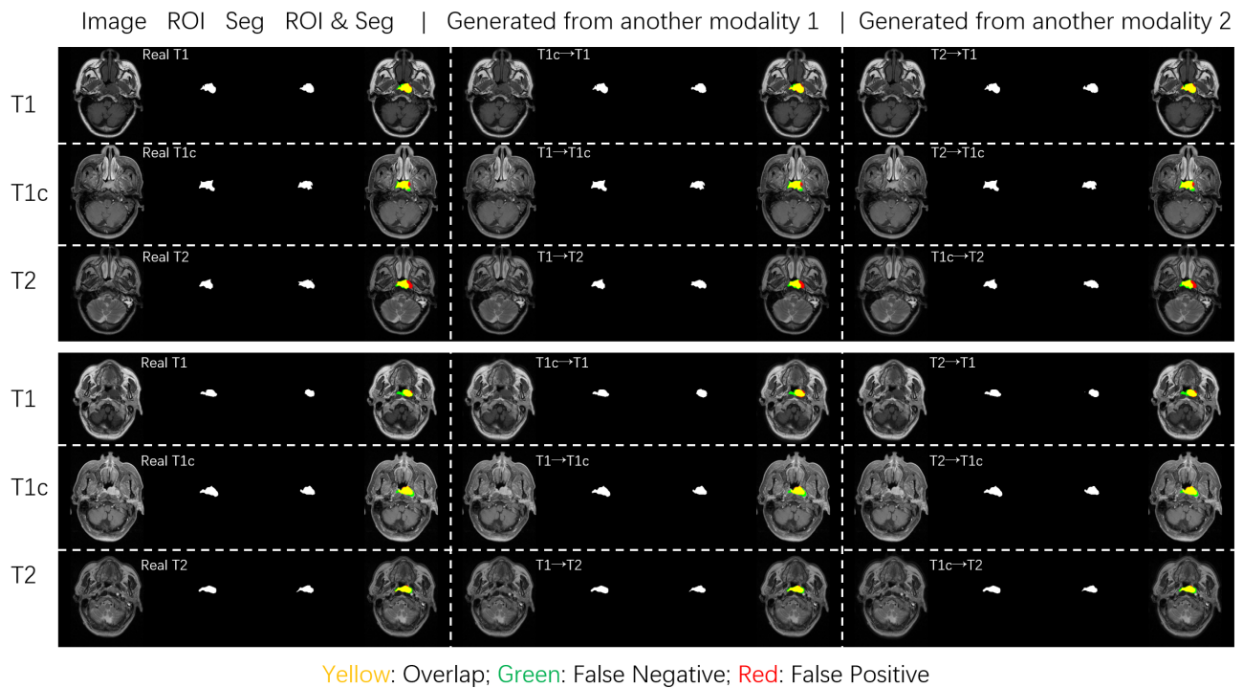

**Figure 9.** NPC segmentation visualization. Every 4 columns images are in one group including input image, NPC Region of Interest (ROI) GT, segment results, and overlapped results in images. We picked two cases here as examples.

**Application to brain tissue segmentation**

In addition to NPC segmentation, we further evaluated the synthesized images on brain tissue segmentation using the widely adopted fully automated tool **FSL** [56] and **SynthSeg** [57]. Specifically, **FSL-BET** was first applied for brain extraction, followed by **SynthSeg**, which segments brain MRI into different anatomy structures based solely on intensity statistics. As a result, the Dice similarity coefficient (DSC in **Table 7**) between tissue segmentations derived from the synthesized images and those obtained from the original images directly reflects the anatomical fidelity and tissue-contrast realism of the synthesized images. However, synthesizing T2 images from T1 or T1c remains more challenging because the hyperintense regions in T1 and T1c are substantially larger than those in T2, resulting in pronounced intensity and tissue-contrast discrepancies that are difficult to faithfully model during cross-modality synthesis. Nevertheless, the spatial distributions of the left and right cerebellum white matter, left and right cerebellum cortex, cerebrospinal fluid (CSF), and brainstem (as shown on **Figure 10**) remain highly consistent with those in the original images.

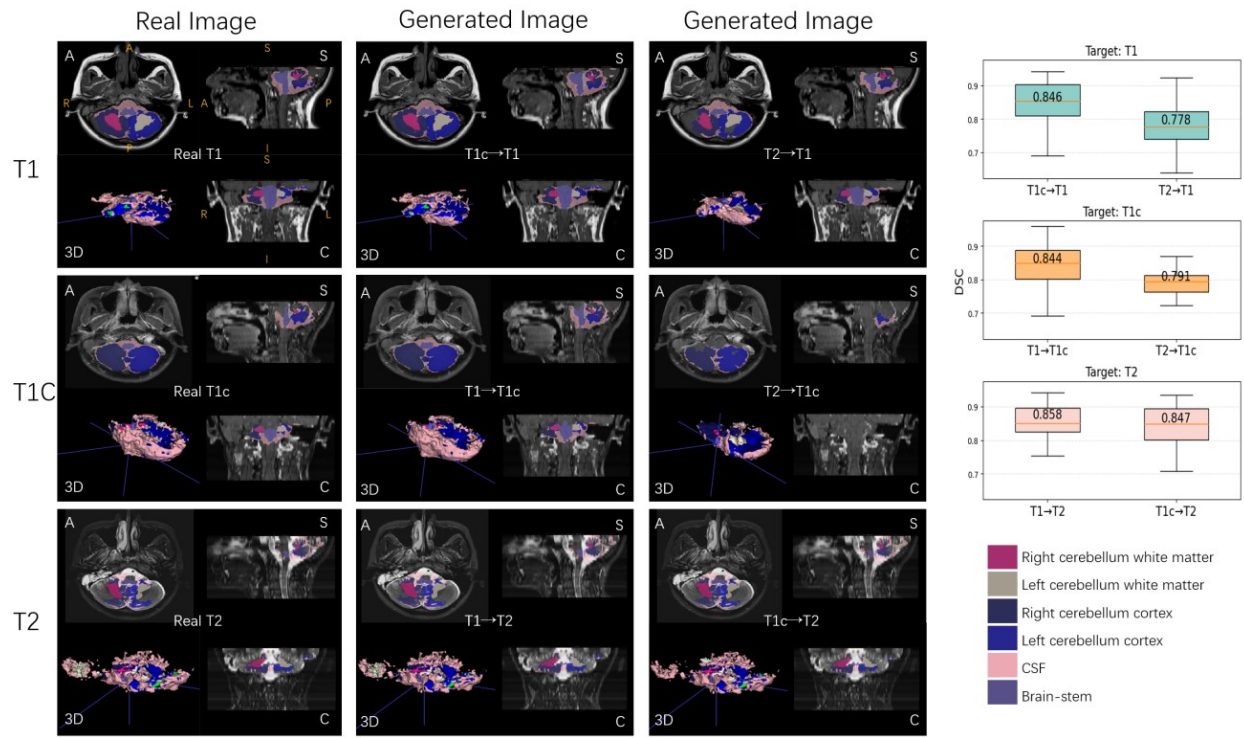

**Figure 10.** Results of brain tissue segmentation using automatic tools. The left panels present representative **A**xial, **S**agittal, and **C**oronal views, together with the corresponding three-dimensional segmentation results, while the right panel shows box plots summarizing the segmentation performance (DSC) across three sets of synthesized images.

**Application to stage prediction.**

To further investigate the clinical utility of the synthesized MRI images, we conducted a downstream task on nasopharyngeal carcinoma (NPC) stage prediction. Specifically, predictive models were trained on the real images from the NPC *External2* dataset and subsequently evaluated on the synthesized images to assess their diagnostic consistency. We focused on four clinically relevant targets, including T stage, N stage, M stage, and overall clinical stage. To characterize the data distribution, six statistical visualizations were generated, comprising five pie charts for T, N, M, clinical stage, and gender, as well as one bar plot illustrating age distribution. Quantitative performance was evaluated using four confusion matrices and four corresponding Receiver Operating Characteristic (ROC) curves. The results demonstrate that models tested on synthesized images achieved comparable predictive behavior, with overall classification Area Under the Curve (AUC) reaching approximately 80% across different staging tasks. Although the classification performance is not too high because the unbalanced distribution of dataset, the findings indicate that the proposed MRI synthesis framework preserves some clinically meaningful features and supports downstream stage prediction, highlighting its potential value for data augmentation and clinical decision support in NPC analysis.

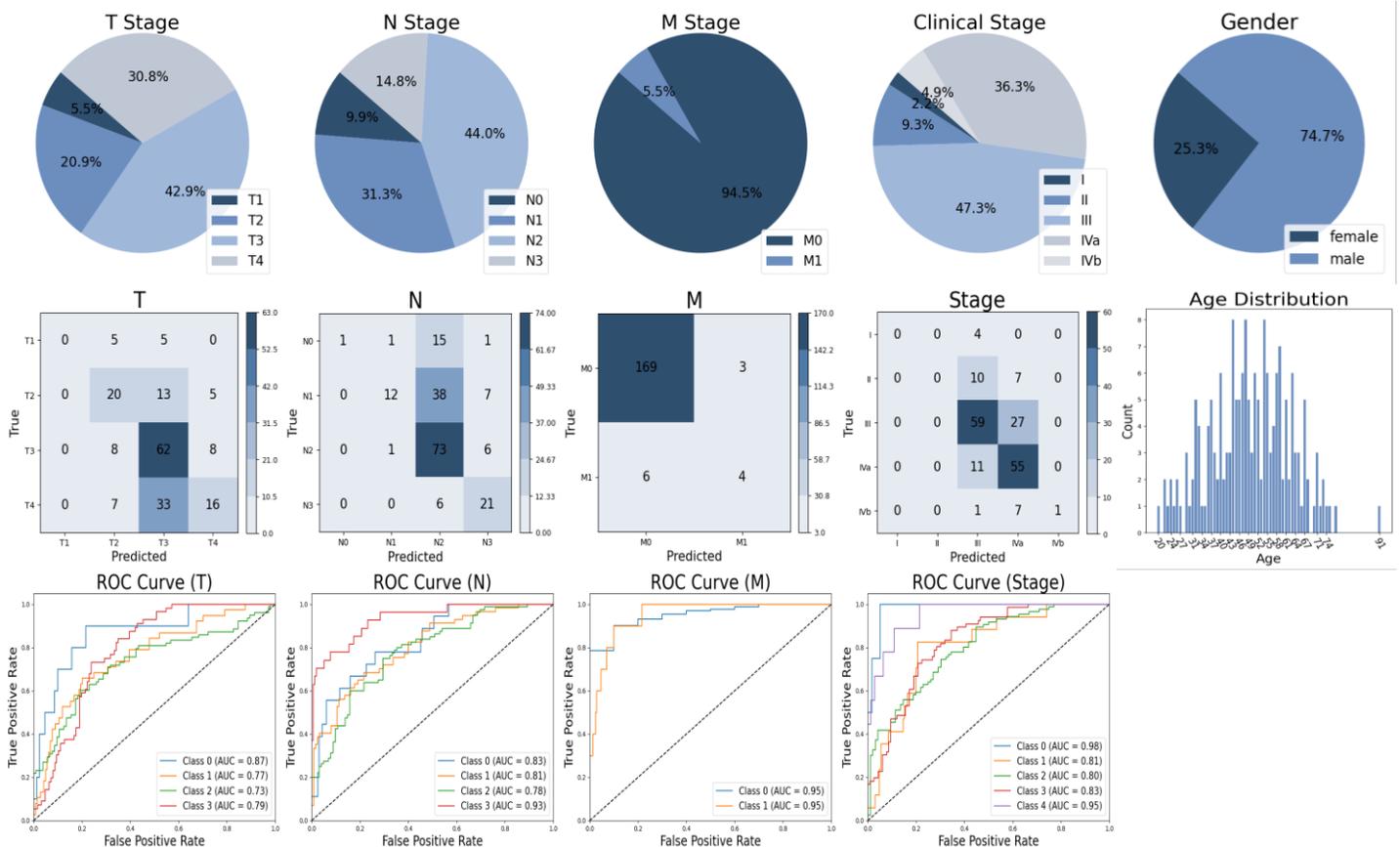

**Figure 11.** Results of stage prediction on synthesized images: 5 pie and bar charts show the distribution of stages, gender and age; confusion matrixes and ROC curves show prediction results.

These three application results collectively show that the synthesized modalities are of sufficient quality for downstream computational analyses, demonstrating the model's strong potential for integration into real clinical workflows where missing or inconsistent modalities often hinder automated processing.

**Outlook**

OmniSyn demonstrates strong synthesis performance and broad applicability across multi-institutional and multi-modality MRI datasets. Nevertheless, several limitations remain, providing important directions for future work. **First**, although OmniSyn demonstrates strong performance across multiple MRI contrasts, the current study focuses on a fixed set of commonly used clinical modalities. The generalization ability of the unified encoder–decoder has not yet been validated on more heterogeneous or rare MRI sequences, such as quantitative parameter maps, advanced diffusion-derived contrasts, or vendor-specific acquisitions. Extending OmniSyn toward a more flexible any-to-any synthesis framework remains an important direction for future work. **Second**, while the training data are multi-institutional, they predominantly consist of images with relatively high diagnostic quality. The behavior of OmniSyn under more challenging conditions—such as severe tumor deformation, post-radiotherapy anatomical changes, or ultra-low-field MRI—has not been systematically evaluated. Future work will explore pathology-enriched or task-specialized training to improve robustness in complex clinical scenarios. **Third**, the current supervision strategy operates primarily at the modality level and does not explicitly enforce fine-grained anatomical or tissue-specific constraints. Although downstream evaluations on NPC tumor segmentation and brain tissue segmentation suggest that the synthesized images preserve clinically relevant structures, incorporating structure-aware supervision or auxiliary anatomical priors may further improve boundary accuracy and pathological detail preservation. **Fourth**, OmniSyn adopts CLIP-based guidance using modality-level textual descriptions, which provides only coarse semantic alignment. More expressive language supervision, such as radiology reports or pathology-specific descriptors, could enable condition-aware synthesis and finer semantic control. In addition, although OmniSyn follows a simple encoder–decoder design with a relatively lightweight parameter count (28.66M) and leverages a fixed ViT-B/32 CLIP model (88M), further analysis of efficiency–performance trade-offs under large-scale deployment remains an open direction. **Overall**, addressing these limitations will substantially expand OmniSyn's capability as a foundation model for cross-modality MRI synthesis and enhance its reliability and clinical relevance across diverse populations and imaging environments.

## Methods

### Study design

This study aims to develop and evaluate a unified foundation model specifically for magnetic resonance imaging (MRI) synthesis in nasopharyngeal carcinoma (NPC), with downstream NPC-related analysis as a primary clinical focus. The overarching goal is to establish a scalable framework that integrates contrastive visual pretraining with language-guided synthesis, enabling a single model to generalize across diverse MRI modalities and clinical settings while maintaining interpretability and adaptability.

As illustrated in **Fig. 1,** OmniSyn consists of **three core components**: a unified visual encoder for modality-invariant representation learning, a CLIP-based vision–language alignment module for semantic guidance, and a lightweight modality-aware decoder for MRI synthesis. **Training** follows a three-stage strategy. First, the unified encoder is pretrained using contrastive learning on multi-institutional MRI data to capture anatomically consistent and modality-agnostic features. Second, CLIP-based vision–language alignment is introduced by training the image–text projection modules, enabling the learned visual representations to align with modality-level textual descriptions. Third, the synthesis decoder is optimized on top of the pretrained and semantically aligned encoder, allowing efficient cross-modality image generation with a compact set of modality-specific parameters.

The **evaluation phase** consisted of comparative experiments, ablation studies, robustness analysis, and downstream clinical tasks. For quantitative comparison, OmniSyn was evaluated on paired and cross-modality MRI synthesis tasks involving T1, T2, and contrast-enhanced T1, and benchmarked against representative baseline methods. Robustness and generalization were assessed using independent multi-institutional datasets acquired from different scanners, reflecting realistic clinical variability. Image fidelity was evaluated using SSIM, PSNR, and MAE. Ablation studies were conducted to analyze the contributions of different architectural designs and training strategies, including variations in the overall framework, CLIP versions, and different strategies. In addition to synthesis quality, downstream NPC-related tasks, including tumor segmentation and brain tissue segmentation, were used to assess the clinical relevance of the synthesized images.

**Unified encoder contrastive learning strategy**

To enable the unified vision encoder to capture both modality-invariant and patient-discriminative representations, we design a two-branch training strategy that jointly enforces **intra-subject consistency** and **inter-subject contrast**. The core idea is to encourage different modalities of the same subject to share a common semantic space while preserving inter-patient uniqueness. For **intra-subject** learning, multi-modal images from the same patient are projected into a shared latent space, where their embeddings are aligned through a combination of global and local constraints. A global vector-level loss based on negative cosine similarity ensures semantic coherence, while a feature map reconstruction loss combining Mean Squared Error (MSE) and L1 loss maintains local anatomical fidelity. For **inter-subject** learning, a contrastive InfoNCE objective is adopted to separate feature distributions of different patients. Each sample is treated as an anchor, with its corresponding modalities as positives and all others in the batch as negatives (**Note**: the batch size should be set as large as possible to keep the enough number of negative samples).

This contrast encourages the encoder to preserve patient-specific diversity while learning generalized anatomical patterns across the population. Formally, the total encoder loss $Loss_{ve}$ is defined as:

$$Loss_{ve} = Loss_{intra} + Loss_{inter} \tag{2}$$

$$Loss_{intra} = Loss_{vector} + Loss_{featuremap} = -similarity + MSELLoss \tag{3}$$

$$Loss_{inter} = Loss_{infoNCE} = \sum_{i \in N} \frac{-1}{|P(i)|} \sum_{p \in P(i)} \log \frac{\exp(z_i \cdot z_p / \tau)}{\sum_{a \in A(i)} 1_{[i \neq a]} \exp(z_i \cdot z_a / \tau)} \tag{4}$$

Here, $z_i$ and $z_p$ denote embeddings of positive pairs (e.g., augmentations of the same slice), while $z_a$ represents negatives (dissimilar slices). The temperature parameter $\tau$ sharpens feature separability. This unified strategy allows the encoder to learn anatomically meaningful, semantically aligned, and generalizable representations across diverse MRI modalities and subjects.

---

**Algorithm 1: Encoder Contrastive Learning**

1: **INPUT:** image of modality i (I_i); text of modality i (T_i); modality_i; CLIP_model. i ∈ {1,2, 3}
2: IF_last_i = IF_i[-1]
3: IF_mid_i = IF_i[:-1]
4: Loss1 = $-\sum_{i \in N} \frac{-1}{|P(i)|} \sum_{p \in P(i)} \log \frac{\exp(z_i \cdot z_p / \tau)}{\sum_{a \in A(i)} 1_{[i \neq a]} \exp(z_i \cdot z_a / \tau)}$, $z_i \in$ IF_last_i
5: Loss2 = $\sum$ MSE(IF_mid_i, IF_mid_j), i≠j
6: **OUTPUT:** Loss1 + Loss2

---

**Vision language alignment strategy**

To incorporate clinical semantic guidance into image synthesis, we employ a contrastive language–image pre-training (CLIP) strategy. While several medical variants such as MedCLIP [32] and RadCLIP [33] have been proposed, we adopted the **original pretrained CLIP** model as our discriminator due to its strong cross-modal alignment capability and generalizable embedding space. CLIP consists of two key components: a **vision encoder** that transforms images into latent visual representations, and a **text encoder** that embeds textual descriptions into the same semantic space. Through contrastive learning, image–text pairs are jointly optimized so that semantically related pairs are drawn closer, while unrelated ones are pushed apart.

However, directly training the vision and text encoders simultaneously in the medical domain often leads to unstable convergence due to limited textual diversity and data imbalance. To address this, we adopted a **two-stage alignment strategy**. In the first stage, the text encoder is fine-tuned using modality labels and structured clinical descriptions, ensuring that text embeddings capture modality-specific semantics. In the second stage, the pretrained text encoder parameters are fixed, and the vision encoder is aligned with these text embeddings by minimizing the cross-modal contrastive loss, effectively linking visual appearance to clinical meaning. This progressive alignment enables the model to learn interpretable visual representations guided by clinical semantics, bridging radiological imaging and textual knowledge.

| **Algorithm 2: CLIP Fine Tuning** |  |
|---|---|
| 1: | **INPUT:** image features of modality i (IF_i). i ∈ {1,2,3} |
| 2: | TF_i = CLIP_model.get_text_feature(T_i) |
| 3: | Loss1 = CrossEntropy(TF_i, modality_i) |
| 4: | Update text encoder |
| 5: | Fix text encoder |
| 6: | IF_i = CLIP_model.get_image_feature(I_i) |
| 7: | Loss2 = CrossEntropy(IF_i, modality_i) |
| 8: | Update image encoder |
| 9: | **OUTPUT:** CLIP_model |

**Unified decoder training strategy**

To achieve coherent and modality-specific MRI synthesis, we design a unified decoder that integrates both visual and semantic conditioning. Specifically, the decoder takes as input two streams of information: (1) **image token** encoded by the unified vision encoder, which provide anatomical and structural priors learned through contrastive pretraining; and (2) **target modality label**, embedded through the simple embedding layer, which convey target label about the desired MRI contrast. By fusing these two modalities, the unified decoder effectively bridges visual understanding and target modality intention, enabling robust cross-contrast image generation.

To optimize the decoder, we employ two complementary loss functions. The first, $Loss_1$, constrains pixel-wise similarity between the synthesized image $\hat{I}$ and the ground truth $I_{gt}$ using a combination of Mean Squared Error (MSE) and L1 loss, ensuring high-fidelity anatomical reconstruction and smooth texture transitions. The second, $Loss_2$, introduces semantic supervision through the CLIP discriminator. We pass the generated images and the target description through the fine-tuned CLIP vision encoder and text encoder separately. Then their corresponding vision/text embeddings differences are computed via **cosine similarity loss**, which measures the discrepancy between predicted and true modality semantics. The final decoder objective is a weighted sum of these two terms, promoting both anatomical accuracy and semantic alignment across modalities.

$$Loss_{pixel} = \text{MSE}(\hat{I}, I_{gt}) + \text{L1}(\hat{I}, I_{gt}) \tag{5}$$

$$Loss_{semantic} = 1 - \frac{E_v \cdot E_t}{\parallel E_v \parallel \parallel E_t \parallel} \tag{6}$$

$$Loss_{vd} = Loss_{pixel} + Loss_{semantic} \tag{7}$$

## Datasets

### Training dataset

The training and validation dataset consisted of **1,077 cases (40,825 slices)** collected from **13 medical institutions** across different regions and vendors, ensuring substantial heterogeneity in acquisition conditions. Participating institutions included Flash Mover MRI Center (*FlashMoverMRI*), Guangyuan First People's Hospital (*GYFPH*), Hong Kong Baptist Hospital (*HKBH*), The University of Hong Kong (*HKU*), Nanfang Hospital (*NFYY*), ProCare Medical (*ProCareMRI*), Prince of Wales Hospital (*PWH*), Queen Elizabeth Hospital (*QEH*), Queen Mary Hospital (*QMH*), Shanghai 6th Hospital (*SH6*), St. Paul's Hospital (*StPaul*), Shenzhen Cancer Hospital (*SZCH*), and Tuen Mun Hospital (*TMH*). The MRI data was acquired using scanners from **Philips, GE, and Siemens** with both **1.5T and 3T** field strengths, including representative models such as *Philips Achieva, Ingenia, and ProdivaCX; GE SignaHDxt, SignaVoyager, and OptimaMR450w;* and *Siemens Avanto, Skyra, Prismafit, and BiographmMR.* This multi-vendor composition provides broad coverage of scanner hardware, imaging protocols, and reconstruction pipelines. The large-scale and diverse nature of this dataset promotes model robustness and prevents overfitting to site-specific intensity patterns or noise characteristics. Among the 40,825 slices, 80% were used for training and 20% for validation, ensuring a balanced partition for supervised optimization and model selection.

### Internal test dataset

An internal test set containing **260 cases (7,319 slices)** was independently collected from **13** same private institutions but not overlapping with those used for training or validation. This internal dataset covered a wide variety of imaging devices and acquisition protocols, with original clinical data. It was used to evaluate the generalization performance of our model under seen clinical environments that share similar regional characteristics but differ in patient cases. The internal test ensures that the model's performance is not biased toward any specific institution, scanner used, or cases during training.

### External1 test dataset (private dataset)

To further evaluate model generalizability, an **external test dataset (External1)** comprising **98 cases (3,331 slices)** was gathered from **independent private centers** distinct from neither the training nor internal test institutions. This dataset represents an unseen population with different acquisition workflows, scanner configurations, and vendor-specific reconstruction characteristics. The 25 institutions included Hong Kong Adventist Hospital (*Adventist*), Alice Ho Miu Ling Nethersole Hospital (*AliceHoMiuLing*), APEX MRI CENTRE (*ApexMRICenter*), Alpha Medical Diagnostic and Laboratory Center (*AplhaMed*), Axon Scanning Centre Limited (*Axon*), Causeway Bay MRI Centre (*Causeway*), Centrum/Centra MRI Centre Limited (*CentralMRI*), Conde S. Januário Hospital (*CHCSJ*), Caritas Medical Centre (*CMC*), EXACT MRI & Diagnostic Centre (*ExactMRI*), Gleneagles Hong Kong Hospital (*Gleneagles*), Hong Kong Advanced Imaging Ltd (*HKAdvanced*), the Hong Kong Integrated Imaging and Endoscopy Diagnostic

Centre (*HKIEDC*), Hong Kong Health Check (*HongKongHealthCheck*), iRad Medical Diagnostic Centre (*iRad*), KPM Healthcare Centre (*KPM*), Mongkok MRI Centre (*MKMRI*), Opus Medical Diagnostic Centre (*OPUS*), Quality HealthCare (*QHDiag*), Hong Kong Sanatorium and Hospital (*Sanatorium*), The Specialists (*Specialists*), St. Teresa's Hospital (*StTeresaHospital*), Trinity Medical Centre (*TrinityMIC*), Union Hospital (*UnionHospital*), United Christian Hospital (*United*), with both synthesized and original clinical data. The inclusion of this external dataset aims to emulate real-world deployment scenarios where the model encounters new clinical sites with potentially unknown imaging variations. The dataset serves as a benchmark for evaluating cross-institutional robustness and the transferability of our foundation model for MRI synthesis. All training, internal test, and external1 test datasets were collected with approval from the Institutional Review Boards of the respective institutions, and written informed consent was obtained from all participants or their parents/legal guardians. Data sharing is restricted due to ethical and institutional regulations.

**External2 test dataset (public dataset)**

An additional **publicly available dataset** [55] containing **182 cases** (5,098 slices, due to missing alignment across the three MRI modalities, we excluded 95 cases from original 277 cases) from **The First People's Hospital of Foshan, China** was adopted as **External2 test dataset** to ensure reproducibility and transparent benchmarking. This dataset, released for research purposes, provides standardized acquisition and labeling protocols and has been widely used in prior studies on MRI modality synthesis and reconstruction. Using this dataset allows for objective comparisons with existing approaches and validation of our framework in a fully open, independent clinical environment. Moreover, this dataset includes not only the MRI image data, but also tumor ROI and clinical data (demographic data, laboratory examination and following up), which could expand evaluation experiments for our foundation model. This public external evaluation thus serves as a critical validation of the model's applicability to external domains under transparent and reproducible settings.

**Table 8.** Dataset statistic information.

| Dataset | Institutions | Cases | Slices | Modalities |
|---|---|---|---|---|
| Train | 13 | 1,077 | 40,825 | |
| Internal Test | 13 | 260 | 7,319 | |
| External1 Test | 25 | 98 | 3,331 | T1, T1c, T2 |
| External2 Test | 1 | 183 | 5,098 | |
| Total | 39 | 1,618 | 56,573 | |

**Construction of modality-specific textual descriptions**

We manually design modality-specific textual descriptions as following to guide the model in capturing modality-aware semantic representations and facilitating effective vision–language alignment. They are designed as prompts for CLIP to match the corresponding labels and then discriminate the generated images.

**Table 9.** Different modality description.

| T1 | "T1-weighted (T1) images provide high-resolution anatomical detail, with fat appearing bright and water appearing dark, useful for visualizing normal tissue structure." |
|---|---|
| T1c | "T1 contrast-enhanced (T1c) images involve the administration of a contrast agent, enhancing vascular structures and providing better visualization of tumors and lesions." |
| T2 | "T2-weighted (T2) images emphasize fluid-rich tissues, with water appearing bright and fat darker, making it ideal for detecting abnormalities like edema or inflammation." |
| FLAIR | "T2 Fluid-Attenuated Inversion Recovery MRI (FLAIR) suppresses cerebrospinal fluid (CSF) signals to better visualize pathological tissues with high water content, such as edema, tumors, or white matter lesions." |
| PD | "Proton density (PD) weighted MRI image highlights tissues with high hydrogen atom concentration, appearing brightest in areas like fat and fluid, while minimizing T1/T2 relaxation effects for enhanced tissue contrast." |
| MRA | "Magnetic Resonance Angiography (MRA) non-invasively images blood vessels by detecting flowing blood signals, aiding in diagnosing vascular abnormalities like stenosis, aneurysms, or malformations." |

**Implementation**

All experiments were implemented using the **PyTorch** deep learning framework, with **SwinUNETR v2** (https://github.com/Project-MONAI/MONAI/blob/dev/monai/networks/nets/swin_unetr.py ) serving as the backbone architecture for both the encoder and decoder components. SwinUNETR v2 follows a hybrid architecture composed of residual CNN blocks, a hierarchical Swin Transformer encoder, a CNN-based decoder, and residual skip connections, which enables effective fusion of local anatomical features and long-range contextual representations in medical images [50]. The original 3D MRI volumes were resampled to an isotropic voxel resolution of $1 \times 1 \times 1$ mm³ to standardize spatial representation across scanners and institutions. Each resampled volume was subsequently resized to $224 \times 224$ pixels per slice to match the network input resolution. The proposed foundation model was trained in three sequential stages—unified vision encoder pretraining, vision–language alignment, and unified vision decoder optimization. The CLIP model, which consists of a dual-encoder architecture with a Vision Transformer (ViT)–based image encoder and a Transformer-based text encoder, was initialized from the pretrained checkpoint "**openai/clip-vit-base-patch32**"(https://huggingface.co/openai/clip-vit-base-patch32), providing both visual and textual embedding spaces for semantic supervision. We employed the Adam optimizer with an initial learning rate of $1\times10^{-4}$, a weight decay of $1\times10^{-5}$, and a cosine annealing schedule to dynamically adjust the learning rate during training. The batch size was set to 24, and mixed-precision training was used to improve computational efficiency. In the encoder stage, the network was optimized using a temperature-scaled InfoNCE contrastive loss and similar loss to enforce intra- and inter-subject feature consistency. During the decoder stage, the total objective combined pixel-wise MSE and L1 losses with a CLIP-based semantic alignment loss, ensuring that synthesized images not only match ground truth contrasts but also align with the correct modality semantics. All training and inference experiments were conducted on Ubuntu 22.04.4 LTS platform with NVIDIA RTX 4090D GPU (24 GB). The average inference time per batch was less than one second. Hyperparameters, including the weighting of loss components, temperature coefficients, and learning-rate schedules, were tuned empirically based on the training subset to achieve optimal synthesis fidelity and generalization.

## Data availability

External 2 test dataset: https://zenodo.org/records/13131827

## Code availability

**Our code:** https://github.com/allenem/omnisyn/ .

**CLIP code:**

CLIP: https://github.com/openai/CLIP,

MedCLIP: https://github.com/RyanWangZf/MedCLIP,

RadCLIP: https://github.com/luzhixiu/RadCLIP,

MediCLIP: https://github.com/cnulab/MediCLIP.

**Networks code:**

UNetr: https://github.com/cameron-cs/unetr-brain-tumour-segmentation/blob/main/model/unetr2d.py,

SwinUnet: https://github.com/HuCaoFighting/Swin-Unet/blob/main/networks/swin_transformer_unet_skip_expand_decoder_sys.py,

SwinUnetr: https://github.com/Project-MONAI/MONAI/blob/dev/monai/networks/nets/swin_unetr.py,

TransUNet: https://github.com/mkara44/transunet_pytorch/blob/main/utils/transunet.py,

ResViT: https://github.com/icon-lab/ResViT/blob/main/models/residual_transformers.py,

BrainMVP: https://github.com/shaohao011/BrainMVP,

TUMSyn: https://github.com/Wangyulin-user/TUMSyn,

## Author information

Yao Pu, Yiming Shi, Zhenxi Zhang, Peixin Yu, Yitao Zhuang, Xiang Wang, Hongzhao Chen, Jing Cai*, Ge Ren*

* Ge Ren is the first corresponding author, and Jing Cai is the second corresponding author. Yao Pu and Ge Ren conceived the study. Yao Pu designed the methodology and implemented the model code. Jing Cai and Ge Ren supervised the project. Yiming Shi and Peixin Yu collected the data and removed redundant information. Yao Pu, Zhenxi Zhang, and Yitao Zhuang designed the downstream tasks and implemented the corresponding code. Xiang Wang and Hongzhao Chen designed the ablation study. All authors reviewed and approved the final manuscript.

**Institutions**: Department of Health Technology and Informatics, The Hong Kong Polytechnic University, Kowloon, Hong Kong SAR, China.